\documentclass[final]{article}

\usepackage{neurips_2025}

\usepackage[utf8]{inputenc} 
\usepackage[T1]{fontenc}    
\usepackage{hyperref}       
\usepackage{url}            
\usepackage{booktabs}       
\usepackage{amsfonts}       
\usepackage{amssymb}        
\usepackage{nicefrac}       
\usepackage{microtype}      
\usepackage{xcolor}         
\usepackage{multirow}
\usepackage{array}
\usepackage{diagbox}
\usepackage{amsmath}
\usepackage{caption}
\usepackage{subcaption}
\usepackage{amssymb}
\usepackage{enumitem}
\usepackage{algorithm}
\usepackage{algcompatible}
\usepackage{hyperref}
\usepackage{url}
\usepackage{makecell}
\usepackage{bm}
\usepackage{graphicx}
\usepackage{enumitem}   
\usepackage{multicol}   
\usepackage{tabularx}
\usepackage{booktabs} 
\usepackage{marvosym}
\usepackage{footmisc}

\setlist[itemize]{topsep=1pt, itemsep=1pt, parsep=0pt}

\title{ValuePilot: A Two-Phase Framework for Value-Driven Decision-Making}

\author{  Yitong Luo$^\dagger$$^{1,2}$,  Ziang Chen$^\dagger$$^{1,2}$, Hou Hei Lam$^{1,2}$ \and \textbf{Jiayu Zhan$^{1,3}$, Junqi Wang$^{1}$, Zhenliang Zhang$^{\text{\Letter}1}$, Xue Feng $^{\text{\Letter}1}$  } \\
\\
$^1$State Key Laboratory of General Artificial Intelligence, BIGAI, \\
$^2$Tsinghua University, 
$^3$Peking University \\
\texttt{chenza22@mails.tsinghua.edu.cn, zlzhang@bigai.ai, fengxue@bigai.ai}
}

\begin{document}

\maketitle
\footnotetext{$^\dagger$Equal contribution. \text{\Letter} Corresponding author.}

\begin{abstract}
  Personalized decision-making is essential for human-AI interaction, enabling AI agents to act in alignment with individual users' value  preferences. As AI systems expand into real-world applications, adapting to personalized values—beyond task completion or collective alignment—has become a critical challenge. We address this by proposing a value-driven approach to personalized decision-making. Human values serve as stable, transferable signals that support consistent and generalizable behavior across contexts. Compared to task-oriented paradigms driven by external rewards and incentives, value-driven decision-making enhances interpretability and enables agents to act appropriately even in novel scenarios. We introduce \textbf{\textit{ValuePilot}}, a two-phase framework consisting of a dataset generation toolkit (DGT) and a decision-making module (DMM). DGT constructs diverse, value-annotated scenarios from a human-LLM collaborative pipeline. DMM learns to evaluate actions based on personal value preferences, enabling context-sensitive, individualized decisions. When evaluated on previously unseen scenarios, DMM outperforms strong LLM baselines—including GPT-5, Claude-Sonnet-4, Gemini-2-flash, and Llama-3.1-70b—in aligning with human action choices. Our results demonstrate that value-driven decision-making is an effective and extensible engineering pathway toward building interpretable, personalized AI agents.

\end{abstract}

\section{Introduction}
Personalized decision-making is a fundamental objective in the development of human-centric AI systems, empowering agents to tailor their behavior based on individual users' goals and value preferences. As artificial intelligence (AI) systems become increasingly embedded in real-world human environments, the demand for such adaptive, interpretable, and human-aligned behavior grows accordingly\citep{lake2017building}. Traditional AI decision-making paradigms, however, remain largely task-oriented: agents are designed to optimize predefined objectives through external rewards\citep{mekni2021artificial, xu2021human}. In contrast, human decisions are shaped not only by goals but also by deeply held personal values. For example, someone who highly values curiosity is more likely to explore unfamiliar environments, even when the task does not explicitly require it. This distinction highlights the need for AI systems that not only complete tasks but also reflect individual value priorities that guide agents' behavior.

The value-driven nature of human decision-making is well-grounded in cognitive science. Theories such as Schwartz's basic human values\citep{schwartz2012overview} and Maslow's hierarchy of needs\citep{maslow1943theory} explain that people act in accordance with internal motivations spanning a range of distinct categories, which we refer to as \textbf{value dimensions} (e.g., curiosity, safety, fairness, intimacy). These value dimensions represent fundamental, abstract principles that guide behavior. Importantly, while these dimensions are relatively stable and shared across individuals, the importance or priority assigned to each dimension can vary significantly from person to person; we term these individualized priorities as \textbf{value preferences}. Decision-making often involves prioritizing or balancing these competing value dimensions to one's unique value preferences. Thus, explicitly modeling both the relevant value dimensions and personalized value preferences enables agents to behave in a human-like manner across a wide variety of circumstances, including scenarios not seen during training.

Despite growing interest in value alignment, current approaches in AI struggle to achieve individualized value-based decision-making. Reinforcement learning from human feedback (RLHF)\citep{christiano2017deep} and direct preference optimization (DPO)\citep{rafailov2024direct} attempt to align agents with human value preferences, but typically rely on aggregated or collective feedback, overlooking inter-individual variability. Structured planning models such as AutoPlan\citep{ouyang2023autoplan} and ReAct\citep{yao2022react} focus on task efficiency without explicitly modeling intrinsic values. These limitations leave two core challenges unresolved: (1) identifying which specific value dimensions are relevant in a given scenario, and (2) selecting actions that align with personalized value preferences while navigating trade-offs across multiple value dimensions.

These challenges are further exacerbated by the lack of suitable datasets. Existing decision-making datasets such as ALFWorld\citep{shridhar2020alfworld} and InterCode\citep{yang2024intercode} are built around extrinsic task completion, offering little insight into how actions relate to underlying human values. Meanwhile, value-related datasets like the World Value Survey (WVS)\citep{inglehart2000world}, Moral Stories\citep{emelin2020moral}, and PAPI\citep{zhu2024personality} either treat human value as a collective whole or lack explicit links between decision scenarios, action options, and fine-grained value dimensions. As a result, they are insufficient for training agents to reason about value-based decisions.

\begin{figure}[ht]
    \centering
    \includegraphics[width=\linewidth]{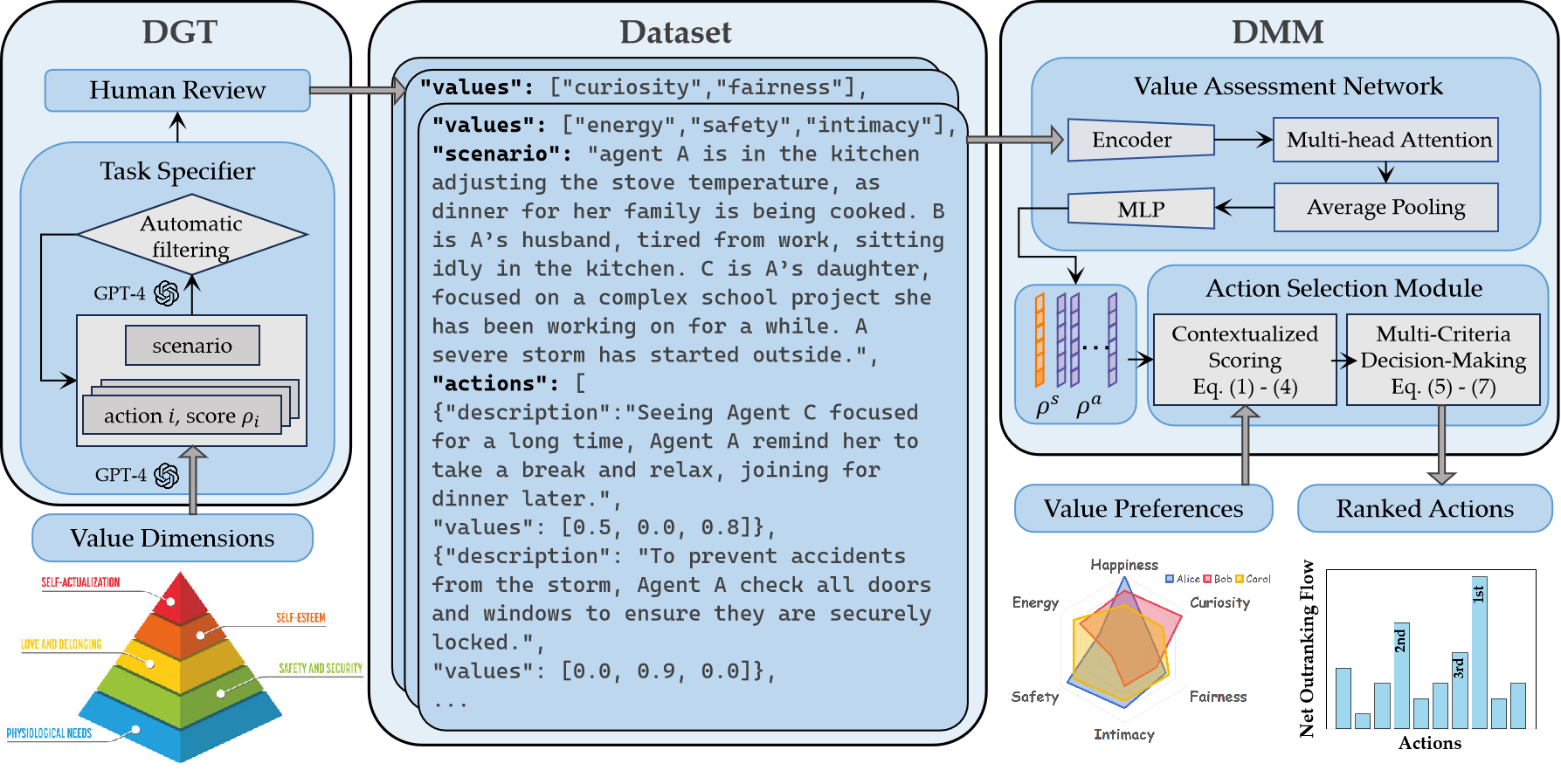}
    \caption{
    A high-level overview of ValuePilot. \textbf{(a) Dataset Generation Toolkit (DGT)}: Given a set of value dimensions, DGT leverages GPT-4 to generate structured decision-making scenarios. Prompts are automatically constructed by the Task Specifier and refined through automated and human-in-the-loop filtering for quality and interpretability.
    \textbf{(b) Dataset}: Each scenario includes the target value dimensions, natural language descriptions of scenarios and candidate actions, and their corresponding value scores ranging from $-1$ to $+1$.
    \textbf{(c) Decision-Making Module (DMM)}: DMM converts the dataset from text descriptions into objective value scores of scenarios, combines them with user-specific value preferences, and then ranks actions using the PROMETHEE method, enabling interpretable, value-consistent action selection.
    }
    \label{fig:pipeline}
\end{figure}

To address these gaps, we propose \textbf{ValuePilot}, a two-phase framework designed to enable personalized, value-driven decision-making in intelligent agents (see \autoref{fig:pipeline}). The framework consists of two key components: a dataset generation toolkit (\textit{DGT}) and a decision-making module (\textit{DMM}). DGT leverages large language models (LLMs) to automatically generate diverse, structured decision-making scenarios, each annotated with numerical scores across multiple value dimensions, thereby providing the data foundation to identify relevant values in context. The generated dataset is further refined through automated and human-in-the-loop filtering for quality and interpretability. DMM then trains on this dataset to recognize the value implications of different actions and integrate them with personalized value preferences, enabling the selection of actions that reflect these individual priorities and adeptly navigate value trade-offs, ultimately generating interpretable action rankings that reflect individual priorities.

We evaluate ValuePilot through extensive experiments across domestic scenarios and human decision-making alignment studies. To assess generalization, we ensure the test scenarios differ from the training set. Results show that ValuePilot significantly outperforms strong LLM baselines (e.g., GPT-4o-mini, Claude-3.5-Sonnet) in aligning with human action choices. These findings demonstrate that value-driven decision-making supports generalizable, interpretable, and personalized behavior, and offer a promising pathway toward building socially-aligned AI systems.

Our contributions are summarized as follows:
\begin{itemize}
    \item We design a multi-stage LLM-based generation pipeline (DGT) that produces structured decision scenarios with interpretable value annotations across diverse value dimensions.
    \item We propose a personalized decision-making module (DMM) that enables explicit value modeling by integrating objective action assessments with subjective user value preferences, leading to interpretable, value-driven decisions in complex scenarios.
    \item By incorporating the PROMETHEE method within the DMM, we enable agents to handle multi-dimensional value trade-offs and generate ranked action lists that reflect human-like decision priorities.
\end{itemize}

\section{Value-Driven Dataset Generation}
To support personalized value-driven decision-making, we require a training dataset that annotates scenarios and actions with structured value information—an element lacking in existing resources. We introduce \textbf{DGT} (Dataset Generation Toolkit), which builds such a dataset through two components: the \textbf{Task Specifier}, which automatically generates and annotates decision scenarios given a predefined set of target value dimensions, and a \textbf{Human Review} module that ensures quality and alignment through manual curation. The resulting dataset encompasses a fine-grained value space inspired by Schwartz’s theory, covering individual (e.g., curiosity), interpersonal (e.g., intimacy), and collective (e.g., fairness) value dimensions. This curated dataset subsequently serves as the training foundation for our Decision-Making Module (DMM), detailed in Section~\ref{sec:section3}. 

\subsection{Task Specifier}
\label{subsec:task_specifier}

The Task Specifier transforms a set of specified value dimensions into structured decision-making scenarios that embed those dimensions in contextually meaningful ways. The module maintains a consistent formatting and scoring schema across all outputs and supports easy integration of new value dimensions, making it extensible to diverse agent tasks and domains. It operates through a three-stage pipeline, designed for extensibility to new value dimensions and diverse agent tasks:

\paragraph{Prompt Construction and Scenario Generation}
Given a set of value dimensions, the system constructs modular prompts for GPT-4 to generate multi-agent domestic scenarios that implicitly encode these value dimensions—for instance, by describing a situation requiring a choice between a familiar safe option and an unfamiliar one to evoke 'safety' and 'curiosity'—
while avoiding direct mention of keywords (e.g., ‘safety’ and ‘curiosity’) to prevent making the value recognition task (see Section~\ref{vare}) too easy.

\paragraph{Action Generation and Value Scoring}
For each scenario, GPT-4 generates 10 diverse possible actions for the protagonist agent in the scenario. To quantify the alignment of each action with the specified value dimensions, we employ a real-valued scoring system. Our quantitative representation method of value weights is inspired by established psychometric methods like the Human Values Scale (HVS), a validated tool for measuring Schwartz’s values which often uses bounded scales to capture varying degrees of importance or endorsement \citep{schwartz1994there, schwartz2007basic}. Similarly, each action in our dataset is accompanied by a score ranging from $-1$ to $+1$ on each dimension:
\begin{itemize}
    \item Values near $-1$ indicate strong contradiction (e.g., a hazardous choice reducing safety).
    \item Values near $0$ reflect neutrality or irrelevance.
    \item Values near $+1$ denote strong alignment (e.g., exploration that promotes curiosity).
\end{itemize}
This bipolar, continuous scaling allows for a nuanced representation of how actions relate to each value.

\paragraph{Automatic Filtering via Re-Evaluation}
To validate value consistency, GPT-4 is re-prompted in a separate session to infer the value dimensions present in each scenario and action, without reference to the original prompt. If the re-identified value dimensions diverge from the target set, the sample is discarded, which improves alignment fidelity.

\subsection{Human Review}

To address risks associated with synthetic data, all generated samples undergo manual review by a four-member team with backgrounds in AI and psychology. Reviewers assess realism, action coherence, scenario diversity, and value alignment. Inconsistent annotations and implausible actions are removed to ensure data quality. This step is critical, as overreliance on unfiltered synthetic data may degrade model robustness over time~\cite{shumailov2024ai}. Human curation ensures the dataset remains accurate, diverse, and suitable for training value-sensitive agents.

\section{Value-Driven Decision Making}

\label{sec:section3}
To enable intelligent agents to make personalized and value-aligned decisions, we propose the 	\textbf{Decision-Making Module (DMM)}—a general framework that integrates scenario understanding with individual value preferences (see \autoref{fig:pipeline}). Trained on our value-annotated dataset, DMM evaluates action alignment with human values through two components: the \textbf{Value Assessment Network}, which estimates how actions impact each value dimension, and the \textbf{Action Selection Module}, which fuses these estimates with user value preferences to produce personalized action rankings. This modular, interpretable structure enables agents to prioritize value-aligned behavior instead of task-based goals.

\subsection{Value Assessment Network}

The Value Assessment Network assesses the alignment between candidate actions and multiple value dimensions within a given scenario. Leveraging a dual-component architecture, it first identifies which values are present and then estimates their degree of alignment for each scenario-action pair. This network operates as the foundational stage of our decision-making pipeline.

We adopt the encoder from T5 (Text-to-Text Transfer Transformer) \citep{raffel2020exploring} to process scenario and action descriptions. As shown in Appendix \ref{appendix:encoder}, we also compare T5 with other encoder models. While T5 slightly outperforms others in accuracy, our framework remains compatible with various encoders, demonstrating its model-agnostic design. The network encodes the scenario and each action into hidden states of shape \( H \times L \times b \), where \( H \) is the hidden dimension, \( L \) is sequence length, and \( b \) is batch size. A multi-head self-attention mechanism (with four heads) is applied to capture semantic relationships between the scenario and actions. The sequence outputs are averaged across \( L \) and passed to a two-layer Multi-Layer Perceptron (MLP) with hidden size 128. The final output is passed through a \texttt{tanh} activation to produce a continuous score in the range $[-1, 1]$ for each value dimension, indicating the degree of positive or negative alignment. These outputs are then passed to the Action Selection Module as \textbf{objective score} of both the scenario and actions for downstream preference-based action selection.

\subsection{Action Selection Module}

Effective decision-making requires accounting for scenario context and user-specific value preferences. The Action Selection Module integrates these through two stages: \textbf{Contextualized Scoring} and \textbf{Multi-Criteria Decision-Making}, with detailed implementation shown in the Appendix \ref{appendix:action_selection}.

\subsubsection{Contextualized Scoring}
\label{subsubsec:contextualized_scoring}
The Contextualized Scoring stage takes two primary inputs: (1) the user's value preference vector $\mathbf{p}$, consisting of self-reported 0-to-1 real-valued scores indicating their emphasis on these value dimensions, and (2) the objective score $\boldsymbol{\rho}$ output by the Value Assessment Network, which quantify the inherent value implications of the scenario and candidate actions. To account for inherent human bias toward moderate scores even when preferences are strong (as described by \textit{Optimal Arousal Theory}\citep{yerkes1908relation}), we apply a sigmoid transformation to the user's raw value preference vector $p$, which emphasizes meaningful preferences for each value dimension $j$ while preserving interpretability.
\begin{equation}
    p'_j = \frac{1}{1 + e^{-(p_j - 0.5) \times 10}}
    \label{eq:1}
\end{equation}
The scaling factor of $10$ ensures that $p_j=0$ is mapped close to $0$ and $p_j=1$ is mapped close to $1$, thereby preserving the extremity of strong preferences. 

For each value dimension \textit{j}, the model uses the \textbf{objective score} $\rho$ (derived from the Value Assessment Network) and computes a \textbf{preference discrepancy score} $d$ for both the scenario ($\rho^s_j$, $d^s_j$) and each candidate action $i$ ($\rho^{a_i}_j$, $d^{a_i}_j$). These quantify alignment with the user’s transformed value preference $p'_j$:
\begin{equation}
    \begin{aligned}
        d^s_j = 1 - \left| |\rho^s_j| - p'_j \right|,\,\,\,\,
        d^{a_i}_j = 1 - \left| |\rho^{a_i}_j| - p'_j \right|
    \end{aligned}
    \label{eq:2}
\end{equation}
Higher $d$ values indicate closer alignment between predicted relevance and user preference, while lower values reflect greater discrepancy.

Integrated scores $r_j^s$ (for scenario) and $r_j^{a_i}$ (for action $i$) on dimension $j$ combine both objective scores and discrepancy scores:
\begin{equation}
    \begin{aligned}
    r^s_j = w d^s_j + (1 - w) \rho^s_j,\,\,\,\,
    r^{a_i}_j = w d^{a_i}_j + (1 - w) \rho^{a_i}_j
    \end{aligned} 
    \label{eq:3}
\end{equation}
Here, \( w \) controls the weight of subjective preference discrepancy relative to objective value scores. Based on conceptual guidance suggesting a \(w\) range of 0.25--0.5 is effective for balancing analytical and subjective factors (detailed in Appendix~\ref{appendix:action_selection}), we consistently utilized \(w=0.3\) throughout our experiments.

To contextualize decisions, each action score $r^{a_i}_j$ is then scaled by the scenario relevance score $r^s_j$, yielding the final score $r_{i,j}$ for action $i$ on dimension $j$:
\begin{equation}
    r_{i,j} = \frac{1}{1 + e^{-\left| r_j^s \right|}} \times r^{a_i}_j 
    \label{eq:4}
\end{equation}
This formulation ensures that actions are weighted not only by their intrinsic value alignment but also by their appropriateness within a given scenario, leading to more interpretable, context-sensitive decision-making. These $r_{i,j}$ scores then serve as the input for the multi-criteria decision-making phase. 

\subsubsection{Multi-Criteria Decision-Making}
\label{subsubsec:mcdm}

We frame action selection as a multi-criteria decision-making (MCDM) problem, as each action must be evaluated across multiple value dimensions. We apply PROMETHEE\citep{brans2016promethee, brans1985note}, a widely used MCDM approach suitable for its outranking principle in handling complex trade-offs, to rank actions via pairwise comparisons across value dimensions. The full rationale for choosing PROMETHEE and its detailed implementation are provided in Appendix \ref{Comparative_experiment} and \autoref{tab:MCDA}. For actions $i$ and $i'$ under dimension $j$, the preference degree of $i$ over $i'$ is: 
\begin{equation}
    V_{ii',j} = \frac{1}{1 + e^{-(r_{i,j} - r_{i',j})}}
    \label{eq:5}
\end{equation}

The overall preference of action $i$ over $i'$, denoted $\Tilde{V}_{ii'}$, is an aggregation weighted by the user's transformed value preferences $p'_j$ across all $m$ value dimensions:
\begin{equation}
    \Tilde{V}_{ii'} = \sum_{j=1}^m p'_j \cdot V_{ii',j} 
    \label{eq:6}
\end{equation}

The final ranking score for each action $i$ is computed as its \textit{net outranking flow} ($\phi_i$). This is defined as its average dominance over other actions ($\phi^+_i$) minus its average weakness ($\phi^-_i$), where $N$ is the total number of candidate actions: 
\begin{equation}
\phi^+_i = \frac{1}{N-1} \sum_{i' \neq i} \Tilde{V}_{ii'}, \quad
\phi^-_i = \frac{1}{N-1} \sum_{i' \neq i} \Tilde{V}_{i'i}, \quad
\phi_i = \phi^+_i - \phi^-_i
\label{eq:7}
\end{equation}
Actions with higher $\phi_i$ are ranked as more preferable, enabling personalized, value-aligned decisions, with the net outranking flow $\phi_i$ providing an interpretable measure of each action's overall preferability.

\section{Experiments}
\subsection{Dataset Preparation}
\label{subsec:dataset_prep}
We select 6 core value dimensions for their importance in daily human life: \textbf{Curiosity}, \textbf{Energy}, \textbf{Security}, \textbf{Happiness}, \textbf{Intimacy}, and \textbf{Fairness}. These were chosen based on relevant literature \citep{maslow1943theory, qiu2022valuenet}, ensuring they are distinct yet representative of a broad range of individual and multi-agent value considerations. After generation via \textit{DGT} and manual screening, our curated dataset comprises \textbf{11,938 scenarios} and \textbf{100,255 actions}. Detailed descriptions of each value dimension, rationale for their selection, and dataset statistics are in Appendix \ref{appendix:dataset}.

\subsection{Value Recognition}\label{vare}
We first assess the performance of ValuePilot’s Value Assessment Network in recognizing inherent value dimensions from scenarios, a foundational capability for value-driven decision-making.

\subsubsection{Baselines and Evaluation Metrics}
We compare our Value Assessment Network against open-source LLMs: Llama-3.5-70b, Llama-3.5-405b, Mixtral-8x22b, and Gemini-1.5-flash. To ensure a fair comparison and enable these LLMs to perform complex value reasoning, we provided them with structured, hierarchical prompts designed to emulate the information processing stages inherent in ValuePilot's pipeline for scenario understanding and value extraction. Performance is assessed using two complementary metrics:
A threshold-based \textbf{Average Accuracy (AvgAcc)}, inspired by Fault Tolerant engineering \citep{lee1990fault}, quantifies classification performance across the six value dimensions. It computes dimension-wise accuracy using absolute error thresholds of \( t = 0.2 \) (practical alignment) and \( t = 0.05 \) (strict alignment). For \( N_t \) test samples and \( D = 6 \) dimensions \( \mathcal{V} = \{v_1,...,v_6\} \), the average accuracy is defined as:
\vspace{-2mm}
\begin{equation}
\text{AvgAcc} = \frac{1}{N_tD} \sum_{i=1}^{N_t}\sum_{d=1}^D \mathbb{I}\big(|\hat{y}_d^{(i)} - y_d^{(i)}| < t\big), \quad
where \,\,\mathbb{I}(\cdot) = 
\begin{cases}
1, & \text{if } |\hat{y}_d^{(i)} - y_d^{(i)}| < t \\
0, & \text{otherwise}
\end{cases}
\end{equation}

This dual-threshold approach addresses both practical and precise alignment needs. Second, \textbf{Mean Absolute Error (MAE)} serves as a continuous metric:
\vspace{-2mm}
\begin{equation}
\text{MAE} = \frac{1}{N_tD}\sum_{i=1}^{N_t}\sum_{d=1}^D |\hat{y}_d^{(i)} - y_d^{(i)}|
\end{equation}
While AvgAcc offers intuitive categorical assessment, MAE proportionally penalizes borderline prediction errors (e.g., predictions of 0.18 vs. 0.22 when $t=0.2$).

\subsubsection{Results}
As shown in \autoref{tab:performance}, our Value Assessment Network outperforms other open-source LLMs across all evaluation metrics. The model achieves 66.70\% average accuracy at \( t = 0.2 \) and 40.00\% at \( t = 0.05 \), representing respective improvements of 15.09 and 14.36 percentage points over the strongest baseline (Gemini-1.5-Flash). Its MAE of 0.19 represents a 36.7\% relative error reduction.

The performance gap widens under stricter thresholds, suggesting our model better captures subtle value distinctions often missed by LLMs. Baseline models exhibit comparable but lower performance (41\% to 52\% accuracy at \( t = 0.2 \)), indicating limitations in their zero-shot value alignment capabilities.

\begin{table}[ht]
\centering
\caption{Comparison between our method and baselines in terms of their ability to recognize inherent values of scenarios. Our method outperforms open-source LLMs.}
\label{tab:performance}
\begin{tabular}{@{}cccc@{}}
\toprule
\textbf{Model}             & \multicolumn{2}{c}{\textbf{AvgAcc (\%)}} & \textbf{MAE} \\ 
                          & $t = 0.2$ & $t = 0.05$ &              \\ \midrule
llama-3.5-70b             & 40.90          & 17.74          & 0.30         \\ 
llama-3.5-405b            & 41.62          & 18.00          & 0.29         \\ 
mixtral-8x22b& 42.71          & 18.39          & 0.29         \\ 
gemini-1.5-flash          & 51.61          & 25.64          & 0.24         \\ 
\multirow{2}{*}{\shortstack{\textbf{Value Assessment} \\ \textbf{Network (Ours)}}}& \multirow{2}{*}{\textbf{66.70}} & \multirow{2}{*}{\textbf{40.00}} & \multirow{2}{*}{\textbf{0.19}}         \\ \\ \bottomrule
\end{tabular}
\end{table}

\subsection{Value-driven Decision-making}
\label{subsec:human_alignment}
To evaluate ValuePilot's DMM in human-like decision-making, we conducted a study designed to capture individualized value preferences and corresponding decision patterns. Specifically, we designed a questionnaire featuring 11 formal domestic scenarios for 40 human subjects. The process began with subjects rating the importance of the six core value dimensions (Curiosity, Energy, Security, Happiness, Intimacy, Fairness) in their daily lives on a 0-to-1 scale. To ensure clarity and familiarize participants with the task structure, a pilot study involving three preliminary questions was first administered. Following this, subjects had the opportunity to rescale their importance ratings for the value dimensions, producing their final, personalized six-dimensional value preference vector. Subsequently, for each of the 11 formal scenarios, participants were presented with a list of candidate actions and asked to rank them according to their willingness to choose each action, guided by their stated value preferences and tendencies. This procedure yielded, for each subject, their unique value preference vector and a set of corresponding ground-truth action rankings for the test scenarios.

These collected individual value preference vectors were then provided as input to our ValuePilot DMM and to the baseline LLMs (Llama-3.1-70b \citep{dubey2024llama}, DeepSeek-R1\citep{guo2025deepseek},  Claude-Sonnet-4 \citep{AnthropicClaudeSonnetWebsite}, Gemini-2.5-flash\citep{comanici2025gemini}, Kimi-K2\citep{team2025kimi}, GPT-4o-mini\citep{hurst2024gpt}, and GPT-5\citep{GPT5TechinicalReport}). For the baseline LLMs, we additionally provide them with detailed definitions of the six value dimensions, along with two real human participants’ value preferences and their action preference rankings in real scenarios as few-shot examples. The models are then tasked with predicting participants’ action preference rankings in real scenarios based on their value preferences. The action rankings generated by each model were subsequently compared against the human-provided rankings to compute alignment similarity. Detailed experimental setups, the full questionnaire, and demographic analysis of the participants are provided in Appendix \ref{appendix:model} and \ref{appendix:human_evaluation}.

\subsubsection{Evaluation Metric}
We use two complementary metrics to comprehensively assess the alignment with human decision patterns. The primary metric, termed \textbf{Order-Sensitive Similarity (OS-Sim)}, extends the Jaccard index through progressive prefix comparison. Let \( i \in \{1, \dots, N_{q}\} \) represent the question sample in the questionnaire. For two sequences of actions \( S^{(i)} = [s_1^{(i)}, s_2^{(i)}, \ldots, s_n^{(i)}] \) and \( T^{(i)} = [t_1^{(i)}, t_2^{(i)}, \ldots, t_n^{(i)}] \), define \( S_d^{(i)} \triangleq \{s_1^{(i)}, \ldots, s_d^{(i)}\} \) and \( T_d^{(i)} \triangleq \{t_1^{(i)}, \ldots, t_d^{(i)}\} \). The similarity at depth \( d \in \{1, \ldots, n\} \) is a modified Jaccard Similarity \(\text{sim}_d^{(i)}(S^{(i)}, T^{(i)}) = \frac{|S_d^{(i)} \cap T_d^{(i)}|}{d}\).   
The mean $\text{OS-Sim}$ is
\vspace{-2.5mm} 
\begin{equation}
\label{eq:os_sim_full} 
\begin{aligned}
    \text{OS-Sim} = \frac{1}{N_q} \sum_{i=1}^{N_q} \text{OS-Sim}^{(i)}, \,\,
    \text{where} \,\, \text{OS-Sim}^{(i)}(S^{(i)}, T^{(i)}) = \frac{1}{n} \sum_{d=1}^n \text{sim}_d^{(i)}.
\end{aligned}
\end{equation}
This design captures both immediate preference alignment in early positions and gradual decision pattern matching throughout the sequence. Detailed examples and the advantages of Os-Sim over methods such as \textit{Spearman Rank Correlation} or \textit{Kendall Tau} are provided in Appendix \ref{appendix:examples_os_sim}. By applying this method, we can objectively compare the ranking lists generated by our model and the LLMs with the actual lists provided by the participants, allowing for a robust evaluation of model performance.

To address real-world constraints where only the primary action selection matters, we introduce \textbf{First-Action Accuracy} as a critical secondary metric. This measure quantifies the percentage of trials where the model's top-ranked action matches the human subject's first choice:
\vspace{-2mm}
\begin{equation}
\text{First-Acc} = \frac{1}{N_{q}}\sum_{i=1}^{N_{q}} \mathbb{I}(s_1^{(i)} = t_1^{(i)}),
\end{equation}
where $\mathbb{I}$ denotes the indicator function. This dual-metric approach evaluates both comprehensive alignment and practical first-choice accuracy.

\subsubsection{Results}
\begin{figure}[!htp]
\centering
\vspace*{-0.5cm} 
\begin{minipage}[t]{0.48\textwidth}
    \centering
    \rule{0pt}{0.5cm} 
    \includegraphics[width=\linewidth]{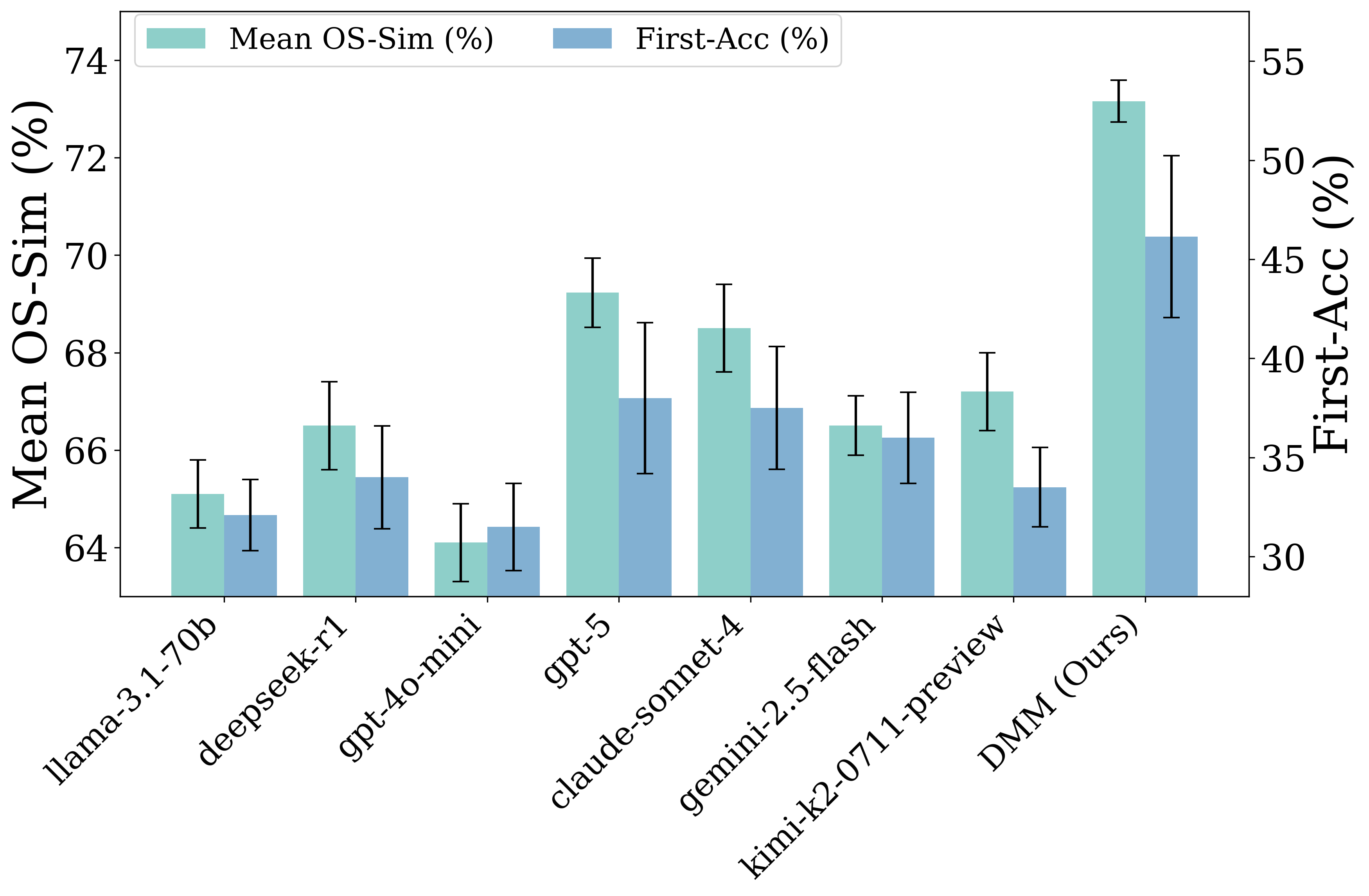}
    \caption{Comparison between our DMM and LLMs in terms of their decision rankings' similarity to human decisions. DMM achieves the highest scores, outperforming the best baseline (GPT-5) by +3.93\% in OS-Sim and +8.13\% in First-Acc.}
    \label{fig:similarity}
\end{minipage}%
\hfill
\begin{minipage}[t]{0.48\textwidth}
    \centering
    \rule{0pt}{0.5cm} 
    \vspace*{1.0cm} 
    \captionof{table}{Ablation Study for DMM.}
    \label{tab:ablation}
    \small 
    \begin{tabular}{@{}ccc@{}}
    \toprule
    \textbf{Model Variant} & \shortstack{\textbf{OS-Sim}\\\textbf{(\%)}} & \shortstack{\textbf{First-Acc}\\\textbf{(\%)}} \\ 
    \midrule
    Only Action & \(60.23 \pm 0.17\) & \(32.27 \pm 0.58\) \\
    w/o Preference & \(61.07 \pm 0.29\) & \(31.82 \pm 0.94\) \\
    w/o Subjective & \(68.93 \pm 0.49\) & \(43.45 \pm 1.16\) \\
    w/o Scenario & \(69.99 \pm 0.57\) & \(43.64 \pm 1.32\) \\
    \textbf{DMM (Full)} & \(\mathbf{73.16 \pm 0.43}\) & \(\mathbf{46.14 \pm 4.09}\) \\
    \bottomrule
    \end{tabular}
\end{minipage}
\end{figure}

As shown in \autoref{fig:similarity}, ValuePilot's DMM significantly outperforms state-of-the-art LLMs in aligning with human decision patterns. Our model achieves a mean OS-Sim of $73.16\% (\pm 0.43\%)$, surpassing the strongest LLM baseline (GPT-5 at $69.23\% \pm 0.71\%$). This highlights its enhanced capability in capturing hierarchical preference structures and nuanced position-sensitive choices.

Notably, in the practically important First-Action Accuracy metric, our DMM achieves $46.14\% (\pm 4.09\%)$, an $8.13\%$ relative improvement over GPT-5 ($38.01\% \pm 3.81\%$). This demonstrates superior fidelity in modeling primary decision determinants, suggesting that explicit value prioritization offers distinct advantages over implicit pattern learning in LLMs.

\subsection{Ablation Study}
We evaluate the full DMM model against four ablated variants, each removing one or all crucial components to analyze its impact.

\begin{itemize}
    \item \textbf{Only Action}: The model ranks actions without considering scenario context or individual value preferences, relying solely on action scores from the Value Assessment Network. Eq.~\eqref{eq:6} changes to \( \Tilde{V_{ii'}} = \sum_{j=1}^{m} \frac{1}{1+e^{-(\rho^{a_i}_j-\rho^{a_i'}_j)}} \).
    \item \textbf{w/o Preference}: The model removes individual value preferences, evaluating actions based purely on their objective scores from the Value Assessment Network. Eq.~\eqref{eq:6} changes to \( \Tilde{V_{ii'}} = \sum_{j=1}^{m} V_{ii',j} \).
    \item \textbf{w/o Subjective}: The model removes subjective preference adjustments, using only objective action and scenario scores without individual bias adaptation. Eq.~\eqref{eq:3} changes to \( r_j = o_j \).
    \item \textbf{w/o Scenario}: The model ranks actions solely based on their individual value impact, without incorporating scenario-based scaling in the final decision. Equation Eq.~\eqref{eq:4} changes to \(r_{i,j} = r_{j}^{a_i} \).
\end{itemize}

Results show that integrating objective action impact, scenario-based adjustments, and individual preference adaptation improves model performance, validating our two-stage framework,

\section{Related Works}

\paragraph{LLMs for Decision-Making and Planning}
LLMs have advanced AI decision-making in interactive environments \citep{nakano2021webgpt, shuster2022blenderbot, glaese2022improving}. Recent methods integrate more explicit reasoning, such as ReAct interleaving reasoning with actions \citep{yao2022react}, or Inner Monologue using feedback for refinement \citep{huang2022inner}, contrasting with affordance-dependent models like SayCan \citep{ahn2022can}. While these, and planning approaches like AutoPlan \citep{ouyang2023autoplan}, enhance LLM agency, they are often task-oriented, with values implicitly tied to task success. They typically lack explicit modeling of fundamental human values or mechanisms for deep personalization of these values, a gap ValuePilot addresses by explicitly incorporating and reasoning over defined value dimensions.

\paragraph{Aligning AI with Individualized Human Values}
Aligning AI with human values is crucial. Reinforcement Learning from Human Feedback (RLHF) \citep{christiano2017deep} and Direct Preference Optimization (DPO) \citep{rafailov2024direct} tune models towards \textit{general} or \textit{collective} human value preferences. However, they often struggle with inter-individual variability and fine-grained value distinctions, generally not learning explicit, interpretable models of different value dimensions. ValuePilot focuses on modeling these explicit dimensions and enabling personalization. Role-Playing Language Agents (RPLAs) \citep{chen2024persona} offer persona-driven interactions, with demographic \citep{huang2023humanity, xu2023expertprompting, gupta2023bias}, character \citep{shao2023character, wang2023rolellm}, and individualized personas \citep{salemi2023lamp, wozniak2024personalized, dalvi2022towards}. While RPLAs advance individualized AI, often enhanced by RAG and long-context memory \citep{chen2024persona}, their focus is typically on conversational fidelity rather than autonomous, explicit value-based action selection in novel scenarios, especially concerning value trade-offs. ValuePilot extends individualization by using its Dataset Generation Toolkit (DGT) to create structured, value-annotated data, and its Decision-Making Module (DMM) to reason over these explicit values and personalized value preferences for action selection.

\paragraph{Multi-Criteria Decision-Making (MCDM) for Value-Driven Choices}
MCDA provides frameworks for balancing conflicting criteria, analogous to an AI balancing diverse human values. Methods like MAUT \citep{dyer2016multiattribute}, AHP \citep{vaidya2006analytic}, and TOPSIS \citep{papathanasiou2018topsis} exist, but PROMETHEE \citep{brans2016promethee, brans1985note} is particularly relevant for its ability to handle distinct criteria by computing net outranking flows. Traditional MCDA applications in AI face challenges with static preference weights and deriving criterion scores dynamically. While recent AI-MCDA integrations aim for adaptability, many remain domain-specific or rely on heuristics. A key challenge is learning action-to-value mappings and incorporating individual-specific trade-offs.
ValuePilot addresses this by using its DGT to enable the DMM's Value Assessment Network to learn objective action-value scores. The DMM then integrates these with user-specific value preferences, employing PROMETHEE for principled resolution of value trade-offs. This combination of learned value assessment and MCDM offers a robust framework for personalized value-driven AI agents.

\section{Conclusion}
In this paper, we presented ValuePilot, a two-phase framework aimed at enhancing personalized decision-making in AI. The framework includes DGT for generating realistic, value-guided scenarios and DMM for ranking actions based on individual value preferences. Through our experiments, DMM demonstrated a strong alignment with human decision-making, outperforming existing LLMs in replicating human choices. We see ValuePilot as an initial step toward the broader goal of value-sensitive AI. While preliminary, our findings suggest the potential of this direction, and we hope it can offer a foundation for future work on socially aligned and interpretable AI agents.



\bibliographystyle{abbrvnat}
\bibliography{custom}

\newpage

\appendix
\section{Dataset}\label{appendix:dataset}

The ValuePilot framework is designed for flexibility, allowing users or system designers to specify the set of value dimensions most relevant to their application domain or research focus. For the experiments presented in this paper, and to demonstrate ValuePilot's capabilities, we selected six core value dimensions to generate our Dataset. This section details the rationale behind this specific selection.

\subsection{Rationale for Selected Value Dimensions}
In understanding human-AI interaction, particularly in everyday environments like the home, identifying the underlying value dynamics that drive choices is paramount. Our work utilizes six core value dimensions: \textbf{Curiosity}, \textbf{Energy}, \textbf{Safety}, \textbf{Happiness}, \textbf{Intimacy}, and \textbf{Fairness}. These dimensions were selected for their relevance to human behavior, drawing inspiration from established psychological frameworks such as Maslow's hierarchy of needs \citep{maslow1943theory} and Schwartz's theory of basic human values \citep{schwartz2012overview}.

\paragraph{Broad Coverage Across Interaction Types}
These six dimensions encompass both single-agent (e.g., an individual's \textit{Curiosity}, \textit{Energy} management, or need for \textit{Safety}) and multi-agent interactions (e.g., seeking \textit{Happiness} or \textit{Intimacy} through social connections, or upholding \textit{Fairness} in group settings \citep{sharma2022human}). This provides a comprehensive lens for analyzing diverse agent behaviors.

\paragraph{Distinctiveness and Minimal Overlap}
While human values are interconnected, the selected dimensions exhibit largely distinct characteristics. For instance, \textit{Curiosity}-driven exploration is not inherently tied to physical \textit{Energy} expenditure, and ensuring environmental \textit{Safety} addresses different needs than fostering social \textit{Intimacy} \citep{sharma2022human}. This conceptual separation aids in clear annotation and modeling.

\paragraph{Alignment with Foundational Theories of Human Motivation}
A meticulous selection process, integrating sociological and psychological perspectives, identified these dimensions as pivotal in shaping human behavior, particularly in domestic contexts.
\begin{itemize}[leftmargin=*] 
    \item \textbf{Curiosity:} Reflects the drive to explore and understand, aligning with self-actualization needs and the value of stimulation \citep{schwartz2012overview, berlyne1966curiosity, berlyne1960conflict}.
    \item \textbf{Energy:} Represents the physiological need for vitality and capacity for activity, fundamental to Maslow's physiological level \citep{maslow1943theory, nix1999revitalization}.
    \item \textbf{Safety:} Pertains to security and stability, a core need in Maslow's hierarchy and a key security value in Schwartz's model \citep{maslow1943theory, schwartz2012overview}.
    \item \textbf{Happiness:} An overarching goal related to well-being and life satisfaction, emerging from the fulfillment of various needs, including esteem and self-actualization \citep{diener2000subjective, lyubomirsky2005pursuing}. 
    \item \textbf{Intimacy:} Fulfills the need for close connection, belonging, and love, crucial for psychological well-being \citep{maslow1943theory, reis2018intimacy}.
    \item \textbf{Fairness:} Relates to justice, equality, and reciprocity in social interactions, aligning with needs for belonging, respect, and universalism values \citep{schwartz2012overview, tyler2015social}.
\end{itemize}

The choice of these six value dimensions aims to span the spectrum of needs described by Maslow, ensuring our analysis is grounded in a well-established psychological framework. They are frequently encountered in everyday domestic life, directly relating to daily living and personal fulfillment. While other values exist, these provide a robust and practical set for modeling prevalent human-agent interaction dynamics. Our goal with this specific set is to demonstrate a pertinent lens for examining and enhancing human-AI symbiosis.

\subsection{Hierarchical Dataset Architecture and Quality Assurance}

The proposed dataset employs a six-tier hierarchical structure (Datasets 1--6) to systematically evaluate computational models across escalating complexity levels of multi-dimensional value interactions. Each tier corresponds to distinct combinatorial configurations of six core value dimensions: \textit{curiosity}, \textit{energy}, \textit{safety}, \textit{happiness}, \textit{intimacy}, and \textit{fairness}. As detailed in \autoref{tab:dataset_stats}, Dataset 1 isolates single-value dynamics (1-D), while subsequent tiers progressively integrate dimensions up to Dataset 6, which encapsulates full six-dimensional (6-D) interactions.

\begin{table*}[htbp]
  \centering
  \caption{Hierarchical Dataset Composition with Filtering Metrics}
  \label{tab:dataset_stats}
  \begin{tabular}{cccccccccc}
    \toprule
    \textbf{Dataset} & 
    \multicolumn{4}{c}{\textbf{Data Filtering}} & 
    \multicolumn{2}{c}{\textbf{Final Scenarios}} & 
    \multicolumn{2}{c}{\textbf{Final Actions}} \\
    \cmidrule(lr){2-5} \cmidrule(lr){6-7} \cmidrule(lr){8-9}
    & 
    \multicolumn{2}{c}{Train} & 
    \multicolumn{2}{c}{Test} & 
    Train & Test & Train & Test \\
    \cmidrule(lr){2-3} \cmidrule(lr){4-5} \cmidrule(lr){6-7} \cmidrule(lr){8-9}
    & 
    Removed & Rate (\%) & 
    Removed & Rate (\%) & 
    & & & \\
    \midrule
    1-D & 2,333 & 12.98 & 430 & 12.08 & 1,771 & 355 & 15,644 & 3,129 \\
    2-D & 2,632 & 15.14 & 430 & 12.25 & 1,689 & 353 & 14,753 & 3,080 \\
    3-D & 3,264 & 18.16 & 501 & 17.20 & 1,704 & 342 & 14,712 & 2,411 \\
    4-D & 3,688 & 22.36 & 728 & 20.59 & 1,609 & 332 & 12,815 & 2,807 \\
    5-D & 3,584 & 23.36 & 863 & 24.38 & 1,412 & 323 & 11,761 & 2,677 \\
    6-D & 4,083 & 22.85 & 901 & 25.17 & 1,714 & 334 & 13,788 & 2,678 \\
    \midrule
    \textbf{Total} & 19,584 & 19.00 & 3,853 & 18.67 & 9,899 & 2,039 & 83,473 & 16,782 \\
    \bottomrule
  \end{tabular}
  \vspace{0.2cm}
\end{table*}

To mitigate potential biases in generated content, we implement a two-stage quality assurance protocol. First, an automated procedure removes value-related pairs explicitly mentioned in input prompts from both scenarios and actions. As quantified in \autoref{tab:dataset_stats}, this eliminates 12.98\%--23.36\% of training actions and 12.08\%--25.17\% of test actions across complexity tiers. Second, the manual phase involves iterative refinement to correct implausible scenarios and inconsistent actions, prioritizing behavioral coherence, interaction plausibility, and dimensional proportionality.

The aggregate statistics validate the stratified design of the dataset, containing 9,899 training scenarios (83,473 actions) and 2,039 test scenarios (16,782 actions) across all tiers. This architecture enables granular analysis of model robustness against dimensional scalability while maintaining methodological rigor through systematic bias mitigation.

\subsection{Dataset Characteristics}\label{appendix:dataset_statistics}
For the dataset we generated, we analyzed the characteristics of the text from multiple perspectives.

\begin{figure}
    \centering
    \includegraphics[width=1.0\linewidth]{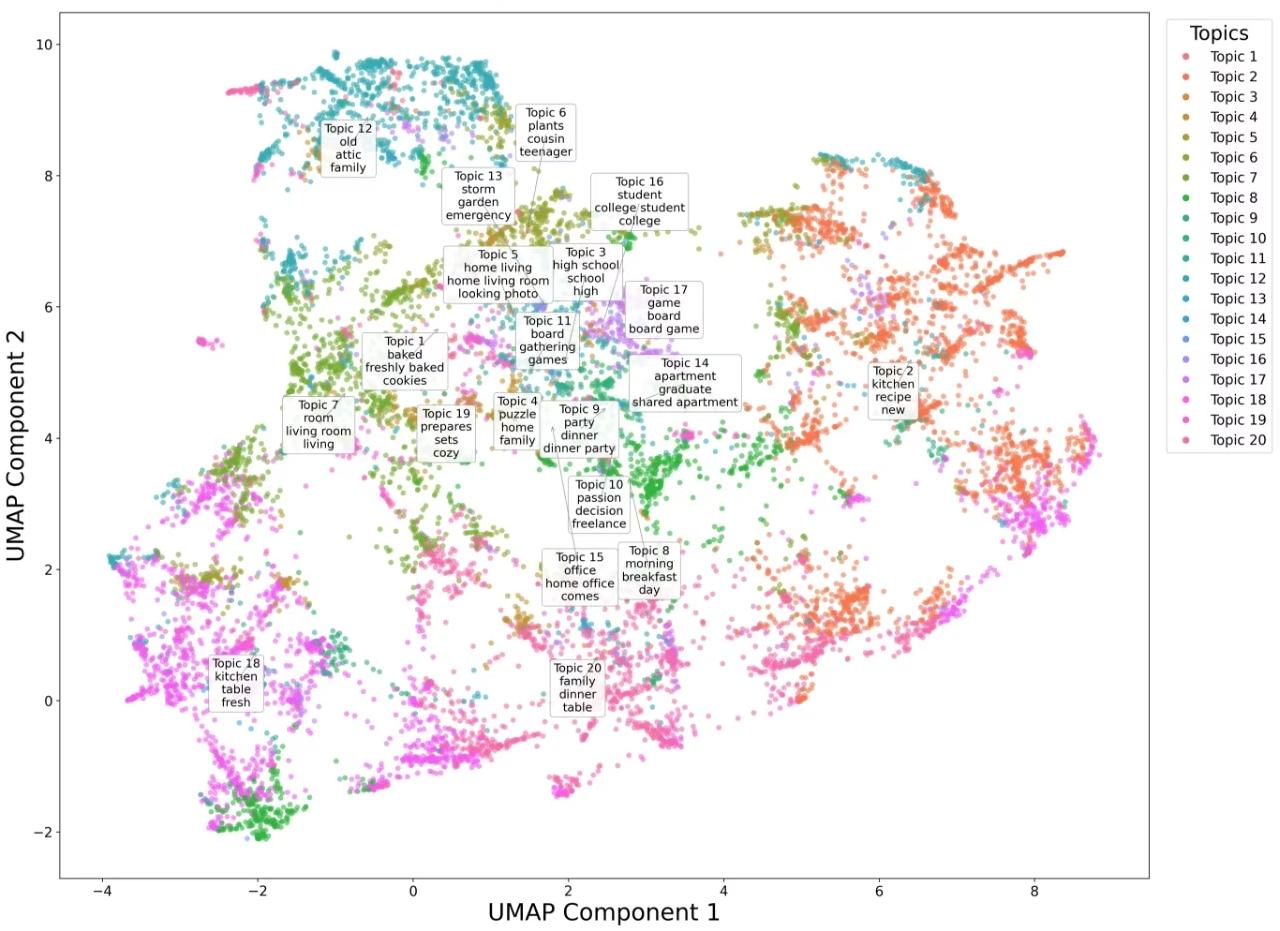}
    \caption{The Topic Space of Scenarios. Using Latent Dirichlet Allocation (LDA), the model automatically groups unlabeled scenario texts into topics based on word co-occurrence patterns. Each cluster is labeled with its dominant topic and annotated with its top three keywords, revealing distinct groupings.}
    \label{fig:ScenarioDiversity}
\end{figure}

\paragraph{Scenario Diversity}(\autoref{fig:ScenarioDiversity})
We extract Term Frequency–Inverse Document Frequency(TF-IDF)\citep{sparck1972statistical, salton1988term} features from scenario texts and apply Latent Dirichlet Allocation (LDA)\citep{blei2003latent} for topic modeling, followed by Uniform Manifold Approximation and Projection (UMAP)\citep{mcinnes2018umap} for dimensionality reduction. The well-separated clusters, labeled by dominant topics, indicate clear categorization, with some overlap reflecting real-world complexity. These results highlight the diversity of scenarios, ranging from everyday activities like cooking to unexpected events such as storms.

\begin{figure*}[ht]
    \centering
    \begin{subfigure}[b]{0.28\textwidth}
        \centering
        \includegraphics[width=\textwidth]{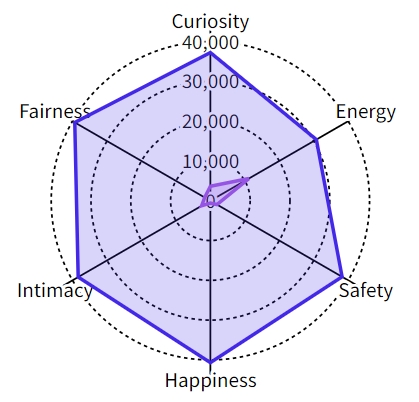}
        \subcaption{Value Distribution in Trainset}
        \label{fig:image1}
    \end{subfigure}
    \hfill
    \begin{subfigure}[b]{0.28\textwidth}
        \centering
        \includegraphics[width=\textwidth]{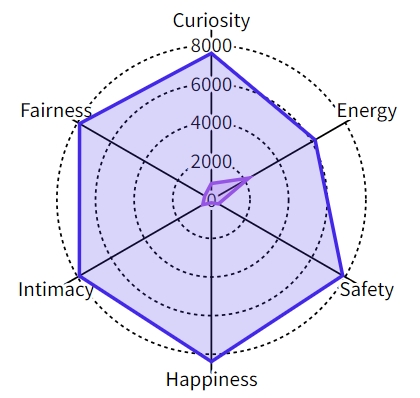}
        \subcaption{Value Distribution in Testset}
        \label{fig:image3}
    \end{subfigure}
    \hfill
    \begin{subfigure}[b]{0.28\textwidth}
        \centering
        \includegraphics[width=\textwidth]{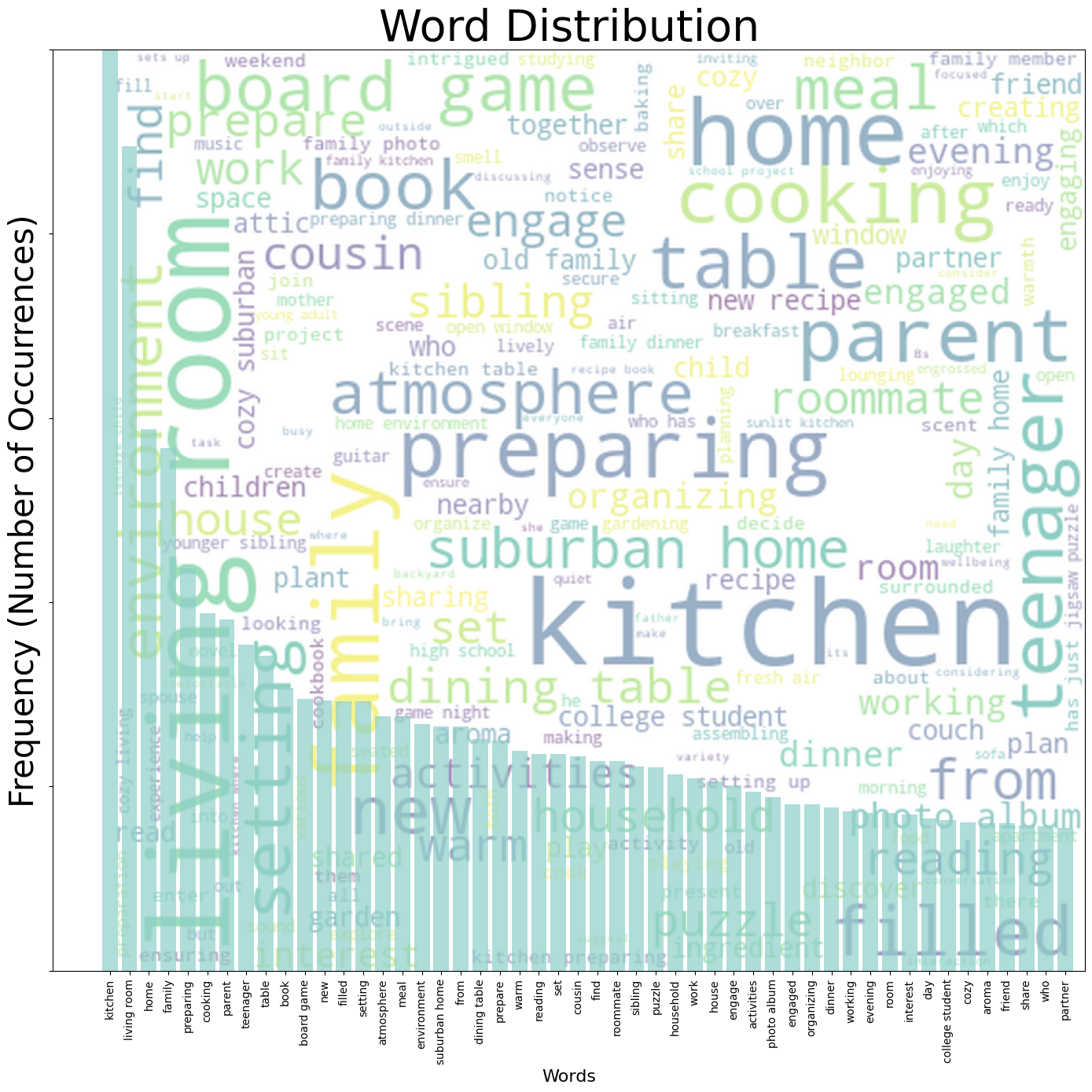}
        \subcaption{Word Frequency Distribution}
        \label{fig:Word Frequency Distribution}
    \end{subfigure}
    \hfill
    \begin{subfigure}[b]{0.28\textwidth}
        \centering
        \includegraphics[width=\textwidth]{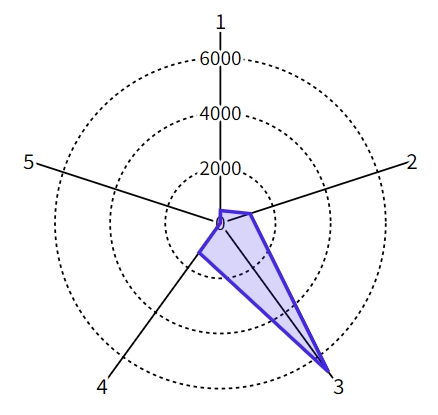}
        \subcaption{Agent Number in Trainset}
        \label{fig:Agent Number in Trainset}
    \end{subfigure}
    \hfill
    \begin{subfigure}[b]{0.28\textwidth}
        \centering
        \includegraphics[width=\textwidth]{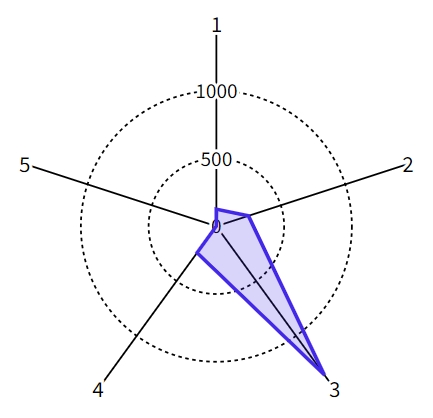}
        \subcaption{Agent Number in Testset}
        \label{fig:Agent Number in Testset}
    \end{subfigure}
    \hfill
    \begin{subfigure}[b]{0.28\textwidth}
        \centering
        \includegraphics[width=\textwidth]{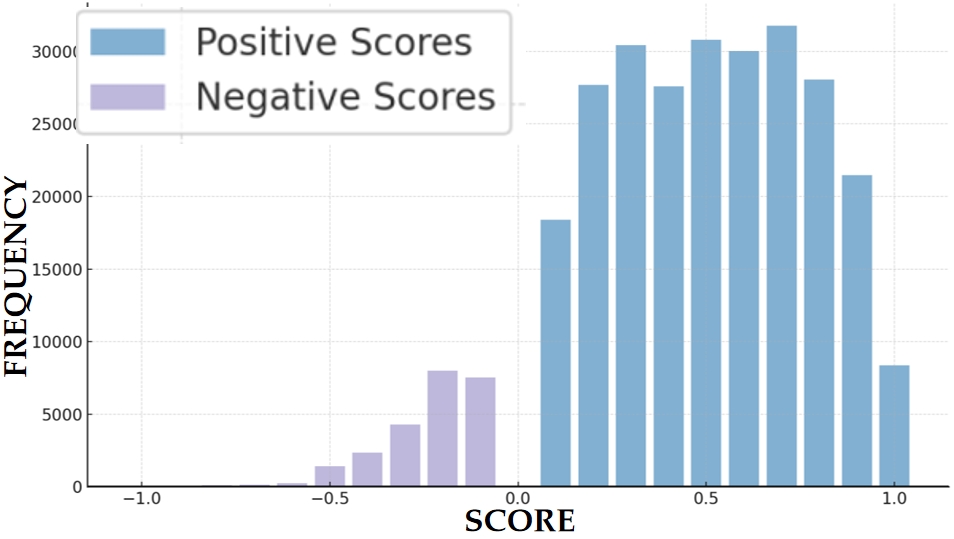}
        \subcaption{Score Frequency Distribution}
        \label{fig:image4}
    \end{subfigure}
    \caption{Statistical characteristics of the dataset; (a) and (b) show the number of times that the scores of all six value dimensions in the dataset appeared as positive and negative values in the scenario; (c) shows the word cloud image of the words that appear in the scenerio; (d) and (e) show the number of agents in the scenarios of the dataset; (f) shows the score distribution of all value dimensions.}
    \label{fig:six_images}
\end{figure*}

\paragraph{Distribution of six-value dimensions}(\autoref{fig:image1} and \autoref{fig:image3}) We counted the number of times that the scores of all six value dimensions in the dataset appeared as positive and negative values in the scenario. Most of the scores of the six value dimensions appeared as positive values, and the total number of the six dimensions is basically the same, though the negative scores of \textit{Energy} are relatively more. The blue box line represents a positive value, and the purple box line represents a negative value. 

\paragraph{Word Frequency Analysis}(\autoref{fig:Word Frequency Distribution}) We performed word frequency analysis on the dataset scenes after removing \textit{stop words}, which are items with less statistical significance such as prepositions, adverbs, quantifiers, etc., and found many interesting phenomena. Among the nouns, \textit{kitchen} and \textit{living room} appear most frequently, representing the places where scenarios often occur. \textit{Preparing} and \textit{cooking} are the most common verbs, representing actions that often occur in the scenario. Notably, \textit{board game} appears frequently. We think that it is reasonable to play board games as a way of multi-person interaction in a home environment that is compatible with happiness, intimacy, and fairness at the same time.

\paragraph{Number of Agents}(\autoref{fig:Agent Number in Trainset} and \autoref{fig:Agent Number in Testset}) We found that the number of agents in the scenarios of the dataset ranges from 1 to 5, and the number of scenarios containing 3 agents is the largest. In the training set and the test set, the corresponding proportion of the number of agents is basically the same.

\paragraph{Score Frequency Distribution}(\autoref{fig:image4}) We calculated the score distribution of all value dimensions in our dataset.

\section{Action Selection Module}\label{appendix:action_selection}
Algorithm~\ref{alg:action-selection} details the procedure used by the Action Selection Module to integrate the Value Assessment Network's objective value scores with personalized user preferences, culminating in a ranked sequence of actions. 

A key parameter in this integration is the weight \(w\), which balances the influence of preference discrepancy scores against objective value scores (as seen in Eq.~\eqref{eq:3} in the main paper). The adjustability of \(w\) allows the system to be tailored for different balances between objective data and subjective alignment, conceptually similar to an LLM's temperature setting. For tasks requiring logical reasoning with some flexibility, a \(w\) range of 0.25--0.5 is generally considered appropriate (see Table~\ref{tab:appendix_w_guidance}). Our empirical testing, which included a grid search and evaluation against human decision patterns, confirmed this, with \(w=0.3\) specifically yielding the best alignment with human choices in our experiments.represents a deliberate balance, giving notable weight to subjective alignment without overshadowing objective assessments. For all experiments presented in this paper, \(w\) was consistently set to $0.3$.

\begin{table}[h]
\centering
\caption{Conceptual Guidance for Subjective Weight (\(w\)) Settings}
\label{tab:appendix_w_guidance}
\begin{tabular}{@{}lll@{}}
\toprule
\textbf{Task Type / Focus} & \textbf{Suggested \(w\) Range} & \textbf{Behavioral Characteristic} \\ \midrule 
Objective Focus        & 0.0--0.25        & Deterministic or fact-based outcomes \\
Analytical Reasoning   & 0.25--0.5        & Logical reasoning with flexibility \\
Balanced Consideration & 0.5--0.75        & Well-rounded, considerate decisions \\
Subjective Emphasis    & 0.75--1.0        & Outcomes strongly driven by personal preference \\ \bottomrule
\end{tabular}
\end{table}

\begin{algorithm}[ht]
\caption{Action Selection Module (DMM)} 
\label{alg:action-selection}
\begin{algorithmic}[1]
\setlength{\jot}{-1pt} 
\raggedright
\Statex \textbf{Input:}
\Statex \quad Scenario Objective Scores: $\boldsymbol{\rho}^s = [\rho^s_1, \ldots, \rho^s_m]$
\Statex \quad Action Objective Scores (for $N$ actions, $m$ values): $\boldsymbol{\rho}^a = [\boldsymbol{\rho}^{a_1}, \ldots, \boldsymbol{\rho}^{a_N}]$, where $\boldsymbol{\rho}^{a_i} = [\rho^{a_i}_1, \ldots, \rho^{a_i}_m]$
\Statex \quad Transformed User Subjective Preferences: $\mathbf{p}' = [p'_1, \ldots, p'_m]$
\Statex \quad Weight for subjective factor: $w$
\Statex \textbf{Parameter:}
\Statex \quad Scenario (text)
\Statex \quad Candidate Actions (list of $N$ actions): $\mathcal{A} = \{action_1, \ldots, action_N\}$
\Statex \quad Value Dimensions (list of $m$ values): $\mathcal{V} = \{value_1, \ldots, value_m\}$
\Statex \textbf{Output:} $RankedActions$
\vspace{1mm} 
\Statex \textbf{\# 1. Contextualized Scoring}
\FOR{$i \in \{1, \ldots, N\}$ (for each action $action_i$)}
    \FOR{$j \in \{1, \ldots, m\}$ (for each value dimension $value_j$)}
        \STATE $d^s_j \leftarrow 1 - \left| |\rho^s_j| - p'_j \right|$; \quad $d^{a_i}_j \leftarrow 1 - \left| |\rho^{a_i}_j| - p'_j \right|$
        \STATE $r^s_j \leftarrow w d^s_j + (1 - w) \rho^s_j$; \quad $r^{a_i}_j \leftarrow w d^{a_i}_j + (1 - w) \rho^{a_i}_j$
        \STATE $r_{i,j} \leftarrow \frac{1}{1 + e^{-|r^s_j|}} \times r^{a_i}_j$ \Comment{Final score for action $i$ on value $j$}
    \ENDFOR
\ENDFOR
\vspace{1mm}
\Statex \textbf{\# 2. Multi-Criteria Decision-Making (PROMETHEE)}
\FOR{$i \in \{1, \ldots, N\}$}
    \FOR{$i' \in \{1, \ldots, N\}$ where $i' \neq i$}
        \STATE $V_{ii',j} \leftarrow \frac{1}{1 + e^{-(r_{i,j} - r_{i',j})}}$ for each $j \in \{1, \ldots, m\}$
        \STATE $\Tilde{V}_{ii'} \leftarrow \sum_{j=1}^{m} p'_j \cdot V_{ii',j}$ \Comment{Aggregated preference of $action_i$ over $action_{i'}$}
    \ENDFOR
\ENDFOR
\vspace{1mm}

\Statex \textbf{\# 3. Calculate Net Outranking Flows}
\FOR{$i \in \{1, \ldots, N\}$}
    \STATE $\phi^+_i \leftarrow \frac{1}{N-1} \sum_{i' \neq i} \Tilde{V}_{ii'}$; \quad $\phi^-_i \leftarrow \frac{1}{N-1} \sum_{i' \neq i} \Tilde{V}_{i'i}$
    \STATE $\phi_i \leftarrow \phi^+_i - \phi^-_i$
\ENDFOR
\vspace{1mm}
\Statex \textbf{\# 4. Rank Actions}
\STATE $RankedActions \leftarrow \text{Sort}(\mathcal{A}, \text{key}=\phi, \text{descending=True})$
\end{algorithmic}
\end{algorithm}

\section{Model Training and Experiment}\label{appendix:model}

\subsection{Implementation Details}
In Experiment 1, we train the networks on a train set comprising 80\% of the total dataset, and then test it on 20\% of the total dataset, aiming to show the efficiency of Value Assessment Network. The experiments are conducted on a system with a single NVIDIA GeForce RTX 4090 GPU (24 GB memory), using CUDA Version 12.6 and driver version 560.70. The models were optimised using the Adam optimizer \citep{kingma2014adam}. The hyperparameters in this optimal configuration consisted of a learning rate range of [1e-3, 1e-5] and a weight decay range of [1e-5, 1e-4] for the Value Assessment Network during Phases 1 and 2. The total training time was approximately 15 minutes.

\subsection{Model Training Procedure}
We use a curriculum learning approach \citep{bengio2009curriculum} to train our models, helping our model build a more robust understanding of basic concepts before tackling more challenging scenarios. To ensure balanced representation, the entire dataset is divided into \( l \) (4 in our experiment) chunks. Each chunk contains a mix of data generated from all combinations of \( n \) (6 in our experiment) value dimensions. The training process follows a predefined order of these groups. This sequence is designed to help the model to adapt and learn efficiently, minimizing the risk of overfitting or underfitting.

Algorithm \ref{alg:training} provides a detailed overview of the curriculum learning approach used to train our models, guiding them through progressively complex data to enhance learning efficiency and performance. In the context of the algorithm you provided, the sequence of dataset $S_{data}$ refers to the predefined order in which different subsets of your data are used during the model training process. This sequence is crucial in the curriculum learning approach because it dictates the order in which the model encounters data of varying difficulty or complexity.

\begin{algorithm}[H]
\caption{Model Training Procedure}
\label{alg:training}
\begin{algorithmic}[1]
\STATE \textbf{Set the sequence of dataset:} $S_{data}= $ [6,1,5,2,4,3]
\STATE \textbf{Initialize} number of chunks to \( l \)

\FOR{each chunk $i$ from 0 to \( l \) - 1}
    \STATE Initialize Datasets
    
    \FOR{each \texttt{n} in training order}
        \STATE \texttt{Combine\_Dataset}.\texttt{concat}(\texttt{Dataset}[\texttt{n}])
        \STATE Initialize early stopping parameters
        
        \FOR{each epoch}
            \STATE Train model, validate model, log metrics, and check for early stopping
            \IF{early stopping criteria are met}
                \STATE Save model
            \ENDIF
        \ENDFOR
    \ENDFOR
\ENDFOR

\end{algorithmic}
\end{algorithm}

\subsection{Additional Results}
\subsubsection{Hyperparameters}
As shown in \autoref{tab:training_results}, Our Networks are trained with different training order and hyperparameters. The results show that a training order of [6, 1, 5, 2, 4, 3] and [lr, wd] sets of {[1e-3, 1e-5], [1e-5, 1e-4]} reaches the highest accuracy and the shortest training time.

\begin{table}
\centering
\setlength{\tabcolsep}{3pt}
\renewcommand{\arraystretch}{0.9}
\small
\begin{tabular}{c>{\centering\arraybackslash}p{0.65cm}>{\centering\arraybackslash}p{0.65cm}>{\centering\arraybackslash}p{0.65cm}>{\centering\arraybackslash}p{0.65cm}|>{\centering\arraybackslash}p{1.2cm}p{1cm}}
\toprule
\multirow{1}{*}{\textbf{Training}} & \multicolumn{2}{c@{}}{\textbf{Phase 1}} & \multicolumn{2}{c@{}}{\textbf{Phase 2}} & \textbf{Time} & \textbf{AvgAcc} \\
\textbf{Order} & {lr} & {wd} & {lr} & {wd} & \textbf{(min:sec)} & \textbf{\,\,(\%)} \\ \midrule
3,4,2,5,1,6 & 1e-4 & 1e-5 & 1e-5 & 1e-4 & 22:43 & 66.06 \\ 
6,1,5,2,4,3 & 1e-4 & 1e-5 & 1e-5 & 1e-4 & 23:25 & 66.15 \\
1,2,3,4,5,6 & 1e-4 & 1e-5 & 1e-5 & 1e-4 & 22:23 & 66.00 \\ 
6,5,4,3,2,1 & 1e-4 & 1e-5 & 1e-5 & 1e-4 & 24:53 & 63.80 \\ 
6,1,5,2,4,3 & 1e-4 & 1e-5 & 1e-4 & 1e-4 & 23:25 & 65.00 \\ 
\textbf{6,1,5,2,4,3} & \textbf{1e-3} & \textbf{1e-5} & \textbf{1e-5} & \textbf{1e-4} & \textbf{14:16} & \textbf{66.70} \\
6,1,5,2,4,3 & 1e-5 & 1e-5 & 1e-5 & 1e-4 & 133:50 & 66.06 \\
\bottomrule
\end{tabular}
\caption{Comparison of Different Training Orders and Hyperparameter Configurations. \textbf{lr} stands for learning rate. \textbf{wd} stands for weight decay. Threshold $t = 0.2$ for AvgAcc.}
\label{tab:training_results}
\vspace{-3pt}
\end{table}

\subsubsection{Encoder Model}\label{appendix:encoder}
To prove that our decision making network works well for a variety of relatively small language models, We replaced the encoder's model, and the experimental results are shown in \autoref{tab:Encoder_selection}.

\begin{table}
    \centering
    \begin{tabular}{cc}
    \toprule 
    \textbf{Encoder}& \textbf{AvgAcc(\%)}\\ 
    \midrule
    \textbf{T5-base}& \textbf{66.70}\\ 
    Flan-T5&  66.00\\ 
    BERT&  65.45\\ 
    RoBERTa&  65.69\\
    \bottomrule
    \end{tabular}
    \caption{Selection study of encoder models. Threshold $t = 0.2$ for Average Accuracy.}
    \label{tab:Encoder_selection}
\end{table}

The T5-base model outputs the highest Average Accuracy. However, upon transforming different encoder models, the final similarity remains relatively consistent, which proves that our fine-tuning method is applicable to different pre-trained language models.

\section{Human Study}\label{appendix:human_evaluation}
\subsection{Demographic Analysis}\label{appendix:Demographic}

The study cohort (N=40) exhibited substantial demographic diversity across age, ethnicity, and socioeconomic dimensions. Subjects spanned adolescents (12.5\%, n=5), young adults (47.5\% aged 18-25), and middle-aged individuals (2.5\% aged 51-60), with balanced gender representation (55.0\% female, 45.0\% male). Educational attainment ranged from secondary education to advanced degrees, with 75\% holding bachelor's qualifications or higher. Occupational diversity spanned 11 distinct professional categories, including academic researchers (15.0\%), technical specialists (7.5\%), and commercial professionals (5.0\%), though students predominated (52.5\%). While the sample is skewed toward younger, educated, urban populations—a common limitation in survey-based studies—this demographic orientation does not substantially compromise the internal validity of findings, though broader generalizability to rural or less-educated groups requires confirmation through targeted sampling in future studies.

\begin{figure*}[ht]
    \centering
    \includegraphics[width=0.95\textwidth]{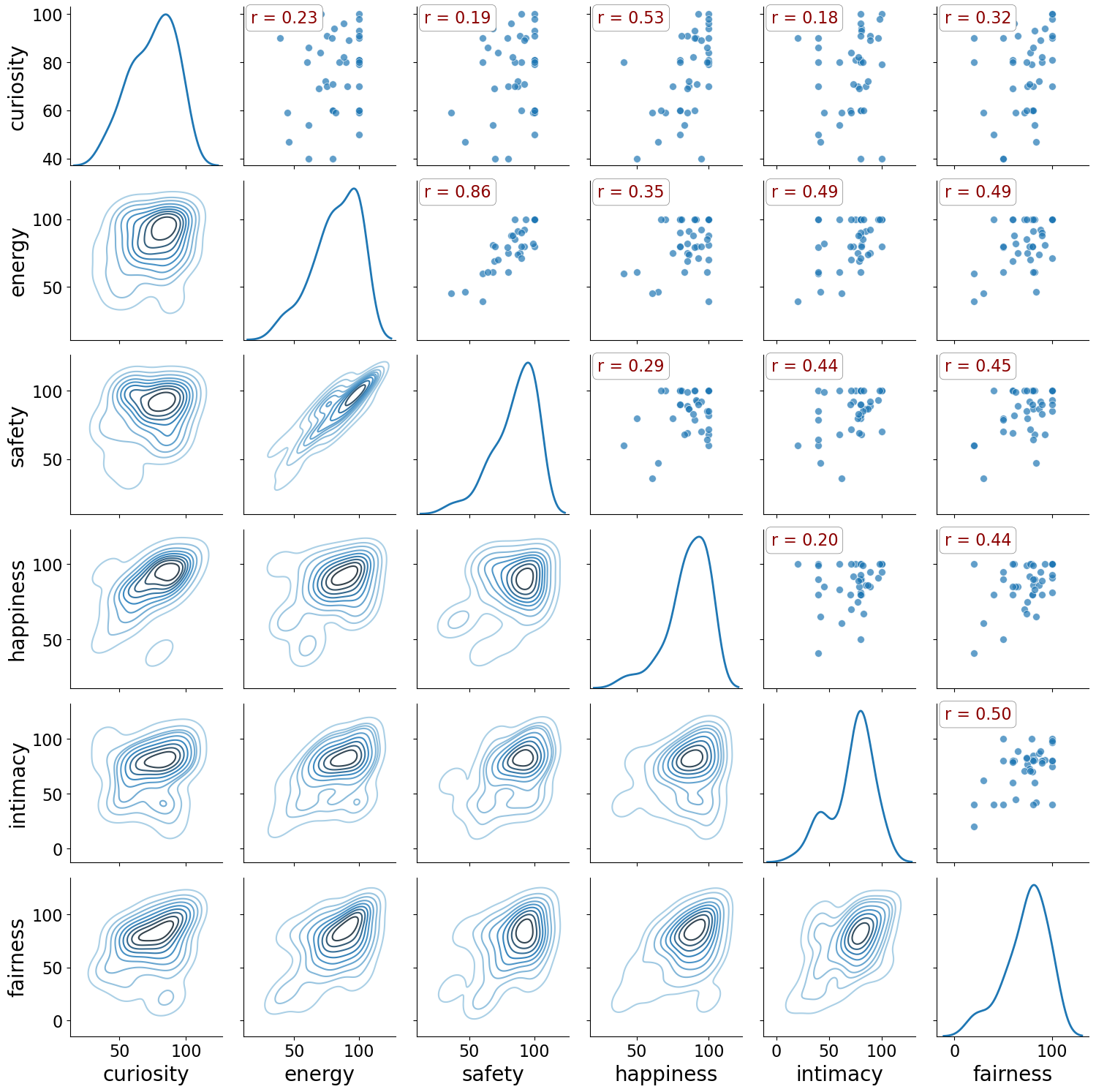}
    \caption{Pairwise Correlations and Distributions of Value Preferences Across the Six Dimensions with Pearson correlation coefficient (r)}
    \label{fig:prefcorr}
\end{figure*}

\begin{figure}[ht]
    \centering
    \includegraphics[width=0.5\columnwidth, keepaspectratio]{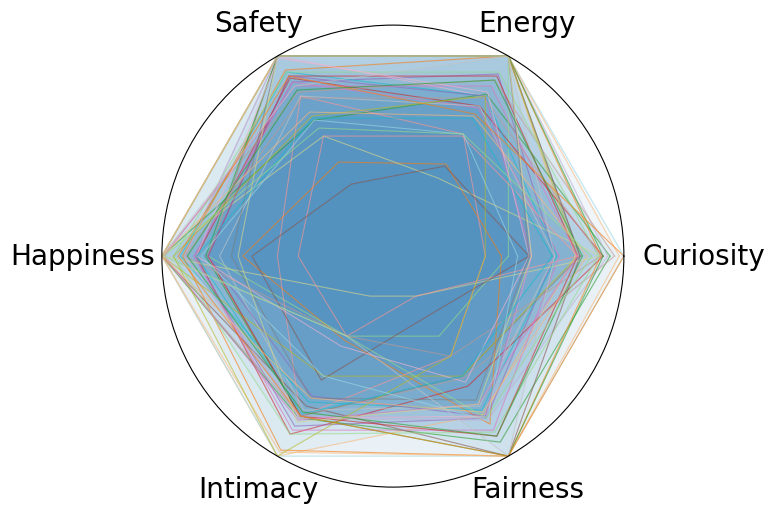}
    \caption{Radar chart showing the value preferences of all subjects. Each polygon represents a subject, and different colors of outline correspond to different subjects.}
    \label{fig:radar}
\end{figure}

In addition to these demographic characteristics, we collected subjects' preference ratings on six value dimensions: \emph{curiosity}, \emph{energy}, \emph{safety}, \emph{happiness}, \emph{intimacy}, and \emph{fairness}. As shown in \autoref{fig:prefcorr} and \autoref{fig:radar}, these preference distributions were notably broad, indicating that our sample captured a wide range of individual differences in value orientations. The pairwise correlation coefficients among the dimensions are also presented, revealing varying degrees of interrelationship across curiosity, energy, safety, happiness, intimacy, and fairness.

\begin{figure}[ht]
    \centering
    \includegraphics[width=0.6\columnwidth,
                    keepaspectratio]{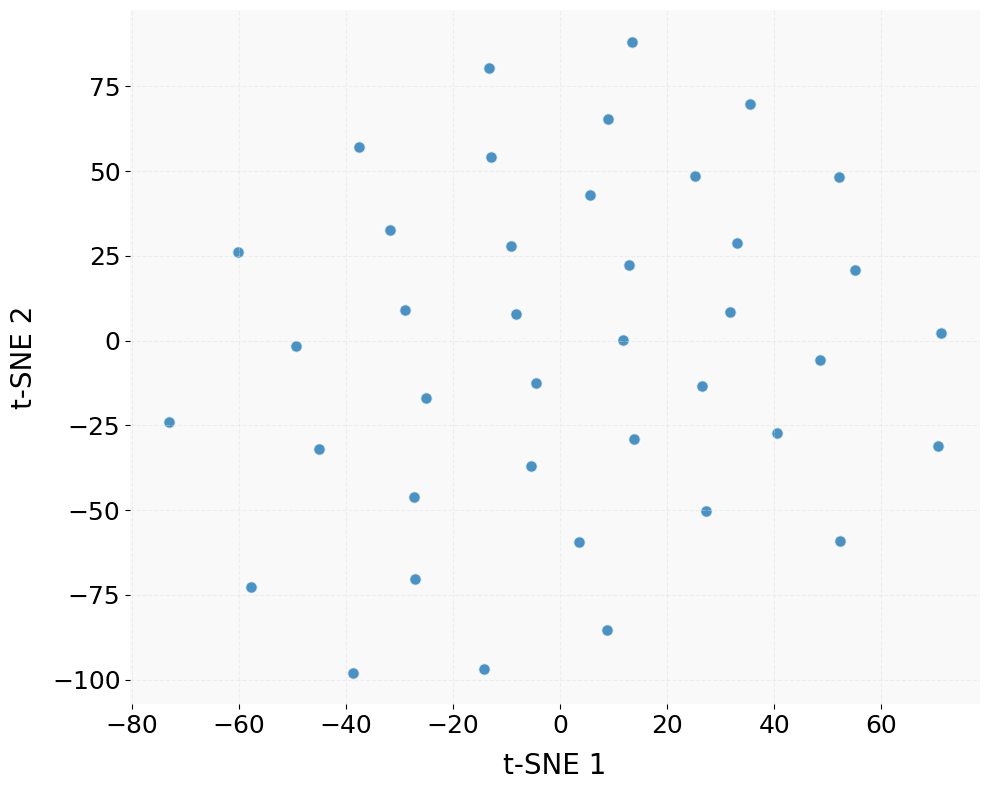}
    \caption{t-SNE Projection of All Subjects' Value Preferences}
    \label{fig:pref_tsne}
\end{figure}

Furthermore, \autoref{fig:pref_tsne} provides a two-dimensional t-SNE visualization of the subjects' preference profiles. The relatively dispersed clustering in this plot suggests that there is no single dominant pattern; instead, subjects' preferences exhibit substantial heterogeneity. Taken together, the breadth of preference distributions, alongside the demographic diversity described earlier, underscores the representativeness of the sample in terms of varying backgrounds and value priorities.

\subsection{Comparative Experiment}\label{Comparative_experiment}
We designed a test questionnaire (see \ref{testquestionnaire}) containing 11 formal test questions. The questionnaire included a detailed introduction to six value dimensions at the beginning, where subjects rated the importance of these dimensions in their daily lives on a scale from 0 to 1, with higher scores indicating greater importance.

To ensure subjects fully understood the experiment's design, a simulation test with three preliminary questions was conducted. This allowed subjects to familiarize themselves with the process, enabling them to provide more accurate data during the formal test. Following the simulation test, subjects re-scored the importance of the six value dimensions, generating a final scoring vector. For each scenario described in the questionnaire, subjects were asked to rank their willingness to choose different actions based on their value dimension scores. This produced a six-dimensional value scoring vector and corresponding action ranking data. Subjects were also asked to identify the value dimensions associated with each action in the scenarios, which facilitated the analysis of the rationality of the data. 

We then input the scenario descriptions and corresponding action lists into our model, along with each subject’s six-dimensional value scoring vector. The model processed this data and outputted action selection rankings based on the subject's value vector. As a comparative experiment, we also input the six-dimensional scores of the subjects into large language models. For each participant, we obtained action selection rankings from both our model and the large language models. We then used the proposed sequence similarity measurement method OS-Sim and First-Acc to compare the ranking lists generated by our model and the large language models with the actual lists provided by the subjects. Then we calculate the Mean OS-Sim, based on the comparative results of different models, fitting the choice of human subjects. Through confidence analysis, we found that the fitting degree of our model to human action selection is better than that of the current mainstream large language model, which proves the effectiveness of our model in analyzing the value dynamics of agents.

We also conducted a experiment to compare the effectiveness of different Multi-Criteria Decision Analysis (MCDA) methods in accurately ranking actions within a given scenario. Based on the experimental results, PROMETHEE shows the highest average accuracy at 73.16\% with a low error margin of ±0.43\%, making it the most reliable method in this scenario, as shown in \autoref{tab:MCDA}.

\begin{table}
\centering
\setlength{\tabcolsep}{4pt}
\begin{tabular}{cc}
\toprule
\textbf{Method} & \textbf{Mean OS-Sim (\%)} \\
\midrule
\textbf{PROMETHEE} & \textbf{73.16 $\pm$ 0.43} \\
AHP & 72.9 $\pm$ 0.56 \\
MAUT & 67.08 $\pm$ 0.47 \\
TOPSIS & 68.19 $\pm$ 0.53 \\
\bottomrule
\end{tabular}
\caption{Comparison of Different Ranking Methods}\label{tab:MCDA}
\vspace{-6pt}
\end{table}

\subsection{Test Questionnaire}
\label{testquestionnaire}
\textbf{Test Instructions:}
We have constructed a value-driven decision-making model for complex environments. The input to the model consists of six value dimensions:
\begin{itemize}
    \item \textbf{Curiosity:}
    An intrinsic desire for knowledge about the world around us. It motivates individuals to actively seek new information and experiences, expanding their understanding of things and enriching their perspectives through exploration. People with high curiosity tend to: actively seek knowledge, bravely explore unknown areas, ask questions to deepen understanding, maintain an open mind, and accept new and diverse views.
    \item \textbf{Energy:}
    The pursuit and satisfaction of physiological energy needs, including food, water, air, sleep, and exercise. It means striving for enough energy to maintain physical and mental health. People with high energy needs tend to: maintain a healthy diet, ensure adequate hydration, secure enough and quality sleep, engage in moderate exercise, and breathe fresh air regularly.
    \item \textbf{Safety:}
    The concern for one's physical and psychological health and the need for stability and predictability in one's environment. This value dimension triggers actions to ensure personal safety, stability in life, and protection from pain, threats, and illnesses, as well as safeguarding personal property. People with high safety needs tend to: maintain physical safety measures, cultivate and maintain mental health, seek stable living conditions, plan for emergencies, and protect personal property.
    \item \textbf{Happiness:}
    A desire for positive emotions and a sense of fulfillment. It drives actions that bring joy, including cultivating positive relationships, achieving personal goals, and enjoying pleasant activities. Happiness manifests in the real world as pleasure and satisfaction with life. People with high happiness tend to: pursue positive emotions, establish meaningful relationships, strive towards personal goals, seek enjoyment and entertainment, and cultivate a sense of accomplishment.
    \item \textbf{Intimacy:}
    Emphasizes deep connections and emotional resonance in various relationships, including family, friendship, romantic, and subordinate relationships. It reflects actions taken to achieve mutual support, understanding, and shared experiences. In the real world, intimate relationships provide warmth and support, promoting individual psychological health and well-being. People with high needs for intimacy tend to: share thoughts, feelings, and experiences, provide support and understanding, willingly share personal privacy, engage in positive and meaningful communication, and pursue a sense of belonging.
    \item \textbf{Fairness:}
    Reflects an individual's pursuit of justice and equality. It motivates actions to create fair and equal conditions in the real world, including actions to ensure fair treatment in social and organizational environments. The goal is to ensure that every individual has equal opportunities and rights. People who highly value fairness tend to: respect equal rights and treatment, promote fair opportunities for everyone, participate in promoting equal decision-making, and treat others justly and fairly.
\end{itemize}
We hope you fill in the questionnaire. We ensure that the questionnaire is filled anonymously and the obtained data is only used to test the generalization effect of the model in the home environment without commercial use.

\textbf{Simulation Test:}
Before addressing specific questions, there is a simulation test to help you understand the specific format of the test.

Please rate your attention to the six value dimensions in your daily life. Ratings are on a scale from 0 to 1, with higher scores indicating greater attention to the value dimension.
\begin{itemize}
    \item \textbf{Curiosity:}\_\_\_\_\_\_\_
    \item \textbf{Energy:}\_\_\_\_\_\_\_
    \item \textbf{Safety:}\_\_\_\_\_\_\_
    \item \textbf{Happiness:}\_\_\_\_\_\_\_
    \item \textbf{Intimacy:}\_\_\_\_\_\_\_
    \item \textbf{Fairness:}\_\_\_\_\_\_\_
\end{itemize}

Below are three simulation test questions below. Please bring the first-person perspective into the following scene. What actions do you tend to choose in this scene?

Please fill in the form below each question. 
The first step, for each action, please give what you think the value dimension of the action is. 
\textbf{For each action, please indicate the value dimensions it represents by placing a checkmark (\checkmark) in the appropriate boxes.}

\vspace{1em} 

\begin{tabular}{|c|c|c|c|c|c|c|c|}
    \hline
    Action & Curiosity & Energy & Safety & Happiness & Intimacy & Fairness & Other Values\\ \hline
    1 & $\checkmark$ &  &  & $\checkmark$ &  &  &  \\ \hline
    2 &  &  &  & $\checkmark$ & $\checkmark$ &  &  \\ \hline
    3 &  & $\checkmark$ &  & $\checkmark$ &  &  &  \\ \hline
    ... & & & & & & & \\ \hline
\end{tabular}

\vspace{1em} 

\textbf{Examples of other value dimensions:} \\
1 Responsibility: Taking the initiative to pick up waste paper falling on the ground and throw it into the trash can may be a test of responsibility.\\
2 Ambitiousness: Choosing to read at home on a good weekend may be out of ambition. \\
3 Authority: Forcing children to do something may be out of the consideration of maintaining their own family authority. 
Etc.

The second step is to prioritize each action in the ' action selection ' according to your willingness to choose, or the possibility of you making the action from high to low. (Blank box to fill in each action before the corresponding serial number 1234...) 
(Note: Different actions are selected for the parallel occurrence relationship in the same scene, and do not arrange in the order of the occurrence time in the scene.)

\begin{center}
\begin{tabular}{|c|c|c|c|c|c|}
    \hline
    Action Preference Sort & 1 & 2 & 3 & 4 & ...\\ \hline
    Action Index &  &  &  &  &  \\ \hline
\end{tabular}
\end{center}

\textbf{Formal Test:}
Presumably, you already know how to test. Here is a formal test with 11 questions.

Please re-score yourself on the importance of the six value dimensions in daily life. Ratings are on a scale from 0 to 1, with higher scores indicating greater attention to the value dimension.

\begin{itemize}
    \item \textbf{Curiosity:}\_\_\_\_\_\_\_
    \item \textbf{Energy:}\_\_\_\_\_\_\_
    \item \textbf{Safety:}\_\_\_\_\_\_\_
    \item \textbf{Happiness:}\_\_\_\_\_\_\_
    \item \textbf{Intimacy:}\_\_\_\_\_\_\_
    \item \textbf{Fairness:}\_\_\_\_\_\_\_
\end{itemize}

Below are 11 formal questions, please finish them in the same requirements as before.

\textbf{Q1:}\\
\textbf{Scenario:} You are a 20-year-old college student. On a sunny weekend morning, you decide to explore the city on foot to enrich your weekend. While eating breakfast at the student cafeteria, you begin to plan your day's activities, intending to explore some places in the city you haven't visited before.

\textbf{Action list:}\\
1. You plan to first go to an online library to collect e-books about the city's development history, so you can have a basic understanding of the city before exploring.\\
2. You are hesitant about exploring unknown corners of the city and think it is better to stick to visiting main and officially recommended attractions to avoid unnecessary trouble.\\
3. You plan to try a mobile app to experience the city's new AR historical guide application, exploring the city in a different way through virtual reality technology.\\
4. You believe that the true joy of hiking adventures lies in discovering the unknown. So, you decide not to make any detailed plans but instead choose a small alley or street you have never walked before, following your intuition to explore the city..
\vspace{0.1cm}

\textbf{Q2:}\\
\textbf{Scenario:} You, B, and C are university roommates. The dorm is usually messy, and no one cleans the common space. Last week, during the routine dormitory cleanliness inspection, you were warned by the dormitory administrator. At this moment, the three of you are seriously discussing how to distribute household chores in the future, with a whiteboard at the door displaying the categories of chores you just listed.

\textbf{Action list:}\\
1. To prevent 'free-riding', you seriously propose to establish a detailed schedule for household chores and assign tasks to each person.\\
2. To prevent omissions, you suggest that everyone think about whether the table listed on the whiteboard fully includes all the chores.\\
3. You think it is unnecessary to assign chores too specifically since you are roommates living together, so you suggest that everyone just do their chores as they go, reminding each other when someone has time.\\
4. You don't want to make things too complicated and propose to decide the assignment of chores by rolling dice.\\
5. You suggest that it's not a big deal and not to be too serious; proposing to go out for dinner together while discussing the distribution of chores.
\vspace{0.1cm}

\textbf{Q3:}\\
\textbf{Scenario:} You, A, and B are a family living together in a suburban area of a quiet town. On the weekend, the three of you are relaxing at home, listening to soft music, enjoying the tranquility of the afternoon. The sun is shining outside, leaves gently swaying, a breeze passing by, creating a relaxed and pleasant atmosphere. You are sitting on the living room sofa, exchanging your moods and recent trivialities.

\textbf{Action list:}\\
1. To enhance family intimacy, you suggest playing a simple riddle game together.\\
2. You feel this weekend should be about relaxation, so you suggest preparing some snacks and watching a light-hearted comedy movie together.\\
3. You think it's a good opportunity to do some housework during the weekend rest time, such as cleaning the living room or organizing clutter, to make the home more tidy and comfortable.\\
4. You suddenly remember a fresh activity and suggest going for a walk in the nearby park together, breathing fresh air and relaxing.\\
5. You suggest using this time to prepare dinner together, trying out some new dishes to enhance family bonding.
\vspace{0.1cm}

\textbf{Q4:}\\
\textbf{Scenario:} You are in the kitchen adjusting the stove temperature, as dinner for your family is being cooked. B is your husband, tired from work, sitting idly in the kitchen. C is your daughter, focused on a complex school project she has been working on for a while. A severe storm has started outside.

\textbf{Action list:}\\
1. To help B relax, you go to talk to him and prepare a hot drink for him.\\
2. To prevent accidents from the storm, you check all doors and windows to ensure they are securely locked.\\
3. To prevent your daughter from being disturbed by the noise of the stove, you close the kitchen door.\\
4. To provide some suggestions and help, you discuss the project with C.\\
5. To create a cozy atmosphere for dinner, you start setting the table.\\
6. Seeing C focused, you remind her to take a break and relax, joining you for dinner later.
\vspace{0.1cm}

\textbf{Q5:}\\
\textbf{Scenario:} You are an enthusiastic middle school student interested in astronomy. B is your brother, a Ph.D. student, and C is your younger sister. It's holiday time, and all three of you are at home. You are casually lying on the living room sofa, flipping through an atlas; B is reading about aerodynamics, and C is preparing a smoothie. An unopened world map puzzle you just bought is on the table, and a computer next to the table displays a news article about breakthroughs in the field of astronomy discussing the possibility of life on an exoplanet.

\textbf{Action list:}\\
1. Looking at the world map on the table, you can't resist opening it and starting to puzzle.\\
2. You propose to C to play a quiz game about the capitals of countries using the atlas, with B acting as the judge.\\
3. Inspired by the article on the computer, you discuss with C whether you believe in the existence of extraterrestrial life.\\
4. On a whim, you build a fort of chairs and blankets on the sofa, imagining it as a spacecraft headed to the planet.
\vspace{0.1cm}

\textbf{Q6:}\\
\textbf{Scenario:} You are a high school student who has just moved to a new community. After school, you sit in your room feeling a bit bored. B, your younger brother, built a sophisticated treehouse with the help of C this morning. He is now happily playing with toys inside the treehouse, which is built very high, taller than your height. C, your neighbor, is a retired carpenter, sitting on his balcony drinking coffee, peacefully watching B play.

\textbf{Action list:}\\
1. To ensure the treehouse is sturdy and your brother is safe, you pick up a toolbox to check for any loose bolts and observe the branches to see if they are strong enough to support the treehouse.\\
2. To alleviate your boredom, you decide to bring some toys and join your brother in the treehouse to play together.\\
3. Curious about the interior structure of the treehouse, you decide to climb up and see how it is built.\\
4. Seeing the exquisitely built treehouse, you become curious about C's career and start a conversation with him.
\vspace{0.1cm}

\textbf{Q7:}\\
\textbf{Scenario:} You and B are college students living together off-campus. One hot summer noon on a weekend, both of you are at home. You go to the kitchen to find something cool to drink, open the fridge, and pour the last bit of orange juice into your cup, eager to drink it to relieve the heat. Just as you are about to finish it, you realize it's the last bottle in the fridge and notice B looking at you.\

\textbf{Action list:}\\
1. You get another cup and pour half the juice for B.\\
2. You plan to pretend you didn't notice it was the last cup, drink it all now, and apologize verbally to B later.\\
3. You plan to drink it all first, then as an apology, go out in the heat later to buy a few more bottles.\\
4. You think it's just a cup of juice and not a big deal, so you drink it all and then go back to your activities.\\
5. You suggest playing rock-paper-scissors with B, where the winner gets the juice, to make the situation fair.\\
6. You stop for a moment to think, then ask B if he wants to drink, and give the juice to him.
\vspace{0.1cm}

\textbf{Q8:}\\
\textbf{Scenario:} You are a teenage boy living in a villa, currently using 3D modeling software in your room to create a digital model of a dinosaur skeleton for a school extracurricular research project, currently facing a bottleneck and unsure how to proceed. B is your father, a paleontologist, sitting at a desk across the room, focused on researching the latest discoveries in his field. C is your pet, a gentle Golden Retriever, comfortably curled up at B's feet on a cozy rug. Near the window in the room, workers are carrying in your family's newly purchased astronomical telescope, getting ready to set it up.

\textbf{Action list:}\\
1. Attracted by the telescope, you go over to join in the excitement, curiously asking the workers about how to operate the telescope.\\
2. Deciding to shift your focus, you go to pet C, covering him with a blanket to keep warm.\\
3. You decide to talk to B about the bottleneck in making the dinosaur model, hoping B can help open your mind.\\
4. Deciding to shift your focus, you take C out to play fetch, hoping to breathe some fresh air and exercise.\\
5. You decide to take a short nap, thinking you might just be too tired, and that waking up might bring new ideas.\\
6. Deciding to shift your focus, you walk around the large house, checking to ensure all windows and doors are secure.
\vspace{0.1cm}

\textbf{Q9:}\\
\textbf{Scenario:} In a cozy family setting on a typical Thursday evening, you, B, and C, three members of the family, are discussing your weekend plans together. You really want to go see a movie, but B and C plan to do DIY cooking and play indoor video games, respectively. You hope to negotiate a consensus to improve everyone's satisfaction with the plan.

\textbf{Action list:}\\
1. You plan to create a schedule for the weekend to see if there's enough time for all activities.\\
2. You insist on going to see the movie and try to convince B and C by explaining how worthwhile the movie is.\\
3. For fairness, you suggest that everyone vote on what to do over the weekend, each person voting for one other activity besides their own.\\
4. To accommodate everyone's ideas, you start thinking about a suitable place that might allow for watching movies, cooking, and playing video games at the same time.\\
5. You suggest that everyone go to their preferred places individually, without the entire family having to stick together.
\vspace{0.1cm}

\textbf{Q10:}\\
\textbf{Scenario:} You are a professional triathlete, just finished a grueling morning training session, and are both tired and hungry. You are thinking about what to eat for a post-training meal while scrolling through your phone. You suddenly come across a news article about a burglary that occurred in your residential area last night, the fourth such incident this month. Your brother B is currently curled up on the sofa watching a live football match.

\textbf{Action list:}\\
1. You decide to grill some lean meat as your meal, considering the protein and nutrition it provides.\\
2. Just finished exercising, you decide to relax first, sitting on the sofa with B to enjoy the live broadcast of the football match and predict which team will win.\\
3. You ask B to wait a moment from watching TV, discussing spending some money on a more advanced security door and adding security cameras.\\
4. Suddenly feeling too tired to do anything, you decide to go to sleep right away to recover your strength.\\

\textbf{Q11:}\\
\textbf{Scenario:} You are a 30-year-old sports enthusiast, currently enjoying a weekend morning with your wife and kids at your comfortable suburban home. While eating brunch, you plan to start a harmonious day and then prepare for a camping trip, discussing the camping plans while eating.
\vspace{0.1cm}

\textbf{Action list:}\\
1. You encourage each family member to drink more water, focusing on their hydration.\\
2. You suggest that the family engage in light stretching activities after eating, to prepare for camping.\\
3. While discussing the details of the camping plans, you ensure every family member has a chance to speak.\\
4. You reminisce about the last time your family went camping, recalling fun and precious moments.\\
5. You delegate the preparation of camping supplies, discussing and deciding on the division of labor fairly with everyone.

\subsection{Examples of OS-Sim}\label{appendix:examples_os_sim}
We demonstrate the unique value of Order-Sensitive Similarity (OS-Sim) through comparative examples contrasting its behavior with traditional rank correlation metrics.

\subsubsection{Example 1: Positional Sensitivity}
Consider two action sequences:
\begin{align*}
    S &= [5, 3, 1, 4, 2] \\
    T &= [3, 1, 5, 4, 2]
\end{align*}

The OS-Sim calculation process reveals progressive alignment:
\[
\begin{array}{c|c|c|c}
d & S_d & T_d & \text{Jaccard}(S_d, T_d) \\
\hline
1 & \{5\} & \{3\} & 0/1 = 0 \\
2 & \{5,3\} & \{3,1\} & 1/2 = 0.5 \\
3 & \{5,3,1\} & \{3,1,5\} & 3/3 = 1 \\
4 & \{5,3,1,4\} & \{3,1,5,4\} & 4/4 = 1 \\
5 & \{5,3,1,4,2\} & \{3,1,5,4,2\} & 5/5 = 1 \\
\end{array}
\]
\[
\text{OS-Sim} = \frac{0 + 0.5 + 1 + 1 + 1}{5} = 0.7
\]

\noindent\textbf{Key Insight}: Although there is full alignment at depths 3-5, the initial mismatches at depths 1-2 reduce the overall score a lot. This reflects real-world scenarios where the ranking of actions in the beginning holds more significance, influencing the outcome more than actions that come later.

\subsubsection{Example 2: Differentiation Capacity}
Consider three recommendation sequences:
\begin{align*}
    A &= [1, 2, 3, 4, 5] \quad \text{(Ground Truth)} \\
    B &= [1, 2, 3, 5, 4] \quad \text{(Back-end Divergence)} \\
    C &= [2, 1, 3, 4, 5] \quad \text{(Front-end Divergence)}
\end{align*}

\begin{table}[h]
    \centering
    \begin{tabular}{l|ccc}
        \toprule
        Metric & $A$ vs $B$ & $A$ vs $C$ & Sensitivity \\
        \midrule
        Spearman's $\rho$ & 0.90 & 0.90 & $\textcolor{red}{\times}$ \\
        Kendall's $\tau$ & 0.80 & 0.80 & $\textcolor{red}{\times}$ \\
        OS-Sim & 0.95 & 0.80 & \textcolor{green}{\checkmark} \\
        \bottomrule
    \end{tabular}
\end{table}

\noindent\textbf{Key Insight}: While traditional rank correlations fail to distinguish between back-end (B) and front-end (C) divergences, OS-Sim successfully quantifies their operational differences through progressive prefix analysis:
\begin{itemize}
    \item \textit{Back-end Divergence}: Less penalty (0.95 vs 1.0) for swapping final elements
    \item \textit{Front-end Divergence}: High penalty (0.80 vs 1.0) from initial misranking
\end{itemize}

This demonstrates OS-Sim's unique capacity to model real-world decision dynamics, where top-ranked actions have higher execution likelihood. Misprioritization of high-stakes actions incurs disproportionately severe consequences compared to lower-priority permutations—a fundamental operational distinction captured by OS-Sim but invisible to conventional sequence-agnostic metrics.

\section{Broader Impacts}\label{appendix:broader-impacts}
The development of our value-driven decision-making framework presents a novel strategy in the creation of AI systems that are not only intelligent but also deeply aligned with human values and ethics. By embedding value dimensions directly into the decision-making process, our approach enhances the ability of AI to make socially responsible and contextually appropriate decisions. This has profound implications for applications in areas like healthcare, education, and personalized services, where the alignment of AI actions with ethical standards and individual preferences is critical. Moreover, our framework's robust capabilities enable AI systems to adapt to a wide range of real-world scenarios, reducing the need for constant retraining and ensuring that these systems remain versatile and effective over time. This adaptability is particularly beneficial in enhancing user interactions, allowing AI to provide more personalized and satisfying experiences. Although potential risks exist, such as biases and privacy concerns associated with misuse of the technology, we believe that the careful design and implementation of our framework, coupled with ongoing vigilance, can mitigate these issues. Ultimately, the positive societal impacts of our work—such as improved decision-making, enhanced personalization, and ethical AI development—far outweigh the challenges, paving the way for AI systems that contribute meaningfully to human well-being and social progress.

\section{Limitations}
\label{app:limitations}

While ValuePilot demonstrates a promising approach to personalized, value-driven AI, we acknowledge certain limitations inherent in our current methodology. The framework primarily emphasizes a novel engineering integration of existing techniques to enable explicit personalized value modeling, rather than on advancing the foundational algorithms for value learning or preference elicitation themselves. Consequently, the current value modeling, while effective for the presented demonstrations, offers a relatively straightforward representation of human values, which are inherently complex, hierarchical, and context-dependent. Furthermore, the method currently employed for eliciting user value preferences relies on direct rating scales. Although practical, this approach may fall short in capturing the nuanced and relative importance of individual values as effectively as more sophisticated psychometric techniques could.


\end{document}